%% file: arxiv.tex
\documentclass[letterpaper]{article} 
\usepackage{aaai2026}  
\usepackage{times}  
\usepackage{helvet}  
\usepackage{courier}  
\usepackage[hyphens]{url}  
\usepackage{graphicx} 
\usepackage{natbib}  
\usepackage{caption} 
\usepackage{algorithm}
\usepackage{algorithmic}
\usepackage{amsmath}
\usepackage{amssymb}
\usepackage{multirow}
\usepackage{xcolor} 
\usepackage{colortbl} 
\usepackage{newfloat}
\usepackage{listings}
\DeclareCaptionStyle{ruled}{labelfont=normalfont,labelsep=colon,strut=off} 
\floatstyle{ruled}
\newfloat{listing}{tb}{lst}{}
\floatname{listing}{Listing}
\title{Transferability of Adversarial Attacks in Video-based MLLMs:\\ A Cross-modal Image-to-Video Approach}

\author{
    Linhao Huang\textsuperscript{\rm 1,\rm 2,\rm 3}\thanks{Equal contribution.}, 
    Xue Jiang\textsuperscript{\rm 3,\rm 4}\footnotemark[1],  
    Zhiqiang Wang\textsuperscript{\rm 5}\footnotemark[1], 
    Wentao Mo\textsuperscript{\rm 1,\rm 3},\\
    Xi Xiao\textsuperscript{\rm 1,\rm 2}\thanks{Corresponding author.}, 
    Yong-Jie Yin\textsuperscript{\rm 6}, 
    Bo Han\textsuperscript{\rm 4}, 
    Feng Zheng\textsuperscript{\rm 3}\footnotemark[2]
}
\affiliations{
    \textsuperscript{\rm 1}Shenzhen International Graduate School, Tsinghua University,\\
    \textsuperscript{\rm 2}Peng Cheng Laboratory, Shenzhen, Guangdong, China\\
    \textsuperscript{\rm 3}Southern University of Science and Technology,\\
    \textsuperscript{\rm 4}TMLR Group, Hong Kong Baptist University,\\
    \textsuperscript{\rm 5}Hong Kong University of Science and Technology,\\
    \textsuperscript{\rm 6}China Electronics Corporation\\
    \{hlh23, mow10\}@mails.tsinghua.edu.cn,
    \{csxjiang, bhanml\}@comp.hkbu.edu.hk,
    zwangmk@connect.ust.hk,
    xiaox@sz.tsinghua.edu.cn,
    yinyongjie@mail.bnu.edu.cn,
    f.zheng@ieee.org
}

\begin{document}

\maketitle

\begin{abstract}
Video-based multimodal large language models (V-MLLMs) have shown vulnerability to adversarial examples in video-text multimodal tasks.
However, the transferability of adversarial videos to unseen models—a common and practical real-world scenario—remains unexplored.  
In this paper, we pioneer an investigation into the transferability of adversarial video samples across V-MLLMs.
We find that existing adversarial attack methods face significant limitations when applied in black-box settings for V-MLLMs, which we attribute to the following shortcomings: (1) lacking generalization in perturbing video features, (2) focusing only on sparse key-frames, and (3) failing to integrate multimodal information.
To address these limitations and deepen the understanding of V-MLLM vulnerabilities in black-box scenarios, we introduce the Image-to-Video MLLM (I2V-MLLM) attack.
In I2V-MLLM, we utilize an image-based multimodal large language model (I-MLLM) as a surrogate model to craft adversarial video samples.
Multimodal interactions and spatiotemporal information are integrated to disrupt video representations within the latent space, improving adversarial transferability.
Additionally, a perturbation propagation technique is introduced to handle different unknown frame sampling strategies.
Experimental results demonstrate that our method can generate adversarial examples that exhibit strong transferability across different V-MLLMs on multiple video-text multimodal tasks. Compared to white-box attacks on these models, our black-box attacks (using BLIP-2 as a surrogate model) achieve competitive performance, with average attack success rate (AASR) of 57.98\% on MSVD-QA and 58.26\% on MSRVTT-QA for Zero-Shot VideoQA tasks, respectively.
\end{abstract}


\begin{figure}[t]
  \centering
  \includegraphics[width=1\linewidth]{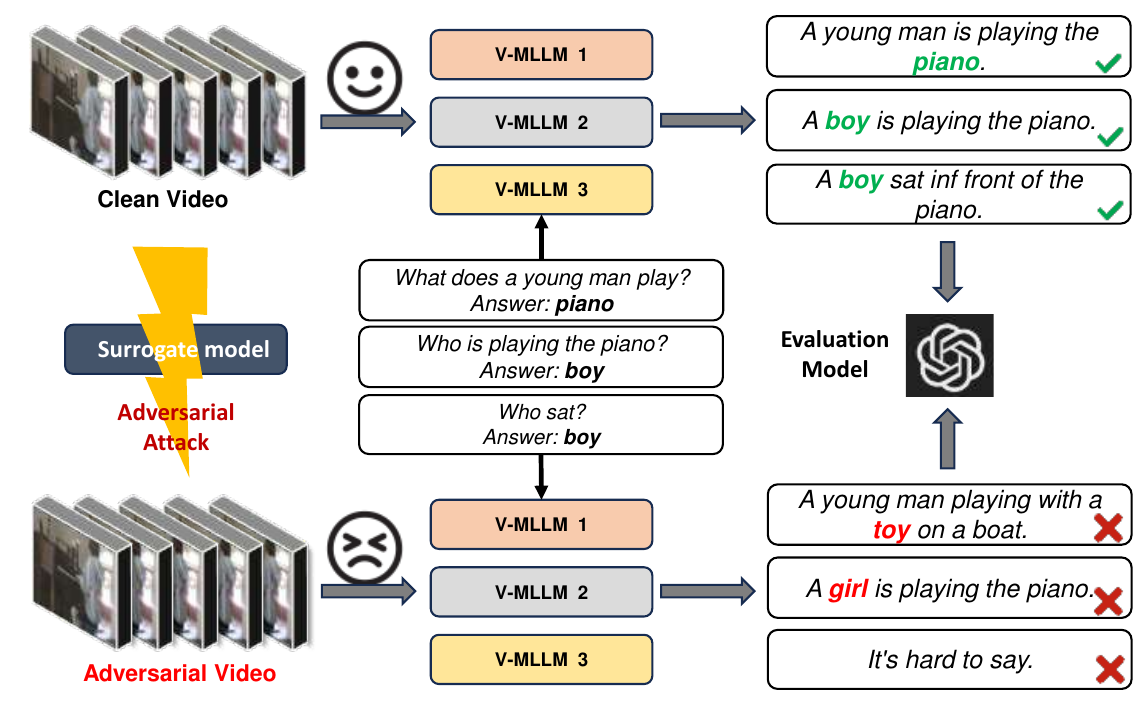} 
  \caption{An example of transferable adversarial attack on different target V-MLLMs for Zero-Shot VideoQA task.}
  \label{fig:0}
  \vspace{-0.5cm}
\end{figure}

\begin{figure*}[t]
  \centering
  \includegraphics[width=1\linewidth]{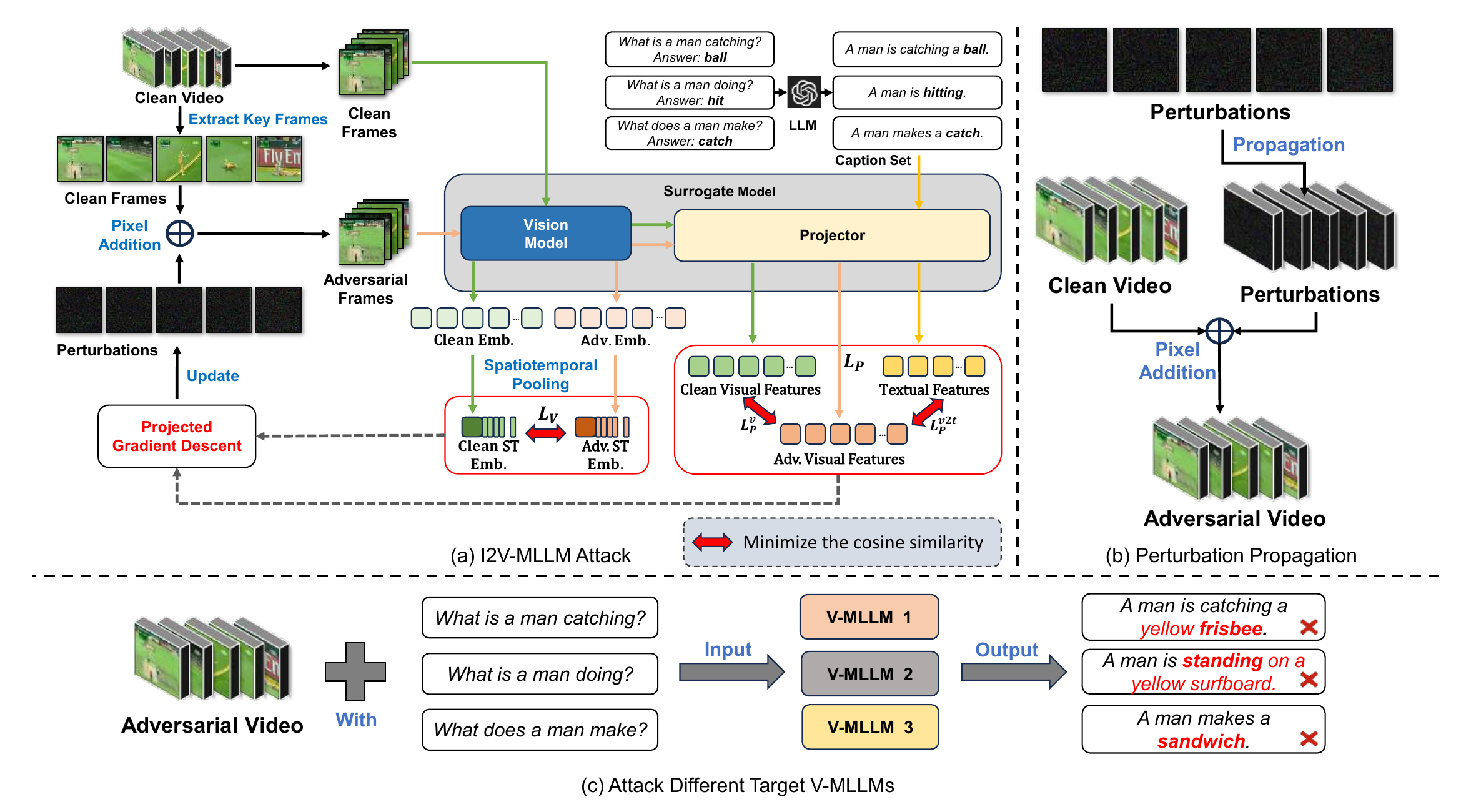} 
  \vspace{-0.5cm}
  \caption{Overview of our proposed method. 
  (a) \textbf{I2V-MLLM Attack:} The clean video is divided into $K$ clips. Key frames are extracted from these clips to form the clean frames $X$, which is then fed into the vision model to extract clean frame-level embeddings $F_V(X)$. These embeddings are subsequently aggregated via spatiotemporal pooling to obtain clean spatiotemporal embeddings $F_V^{st}(X)$. Perturbations are initialized and added to clean frames $X$ to generate adversarial frames $X_{adv}$. The same process is applied to extract $F_V(X_{adv})$ and $F_V^{st}(X_{adv})$. An LLM reformulates the QA pairs into a caption set $T$. $F_V(X)$, $F_V(X_{adv})$, and $T$ are  then passed through the projector to extract visual features $F_P^v(X)$, adversarial visual features $F_P^v(X_{adv})$, and textual features $F_P^t(T)$, respectively. Perturbations are updated via the PGD method by minimizing three cosine similarity-based losses: $L_V$, $L_P^v$, and $L_P^{v2t}$.
  (b) \textbf{Perturbation Propagation:} The final perturbations applied to key-frames are propagated back to their corresponding video clips to construct the adversarial video.
  (c) \textbf{Attack Different Target V-MLLMs.}}
  \label{fig:1}
  \vspace{-0.5cm}
\end{figure*}

\section{Introduction}

Recent work has shown that video-based multimodal large language models (V-MLLMs) are vulnerable to adversarial video samples \cite{li2024fmmattackflowbasedmultimodaladversarial}, even though they have achieved remarkable performance on a wide range of video-text multimodal tasks \cite{li2024videochatchatcentricvideounderstanding, jin2024chatuniviunifiedvisualrepresentation, lin2024videollavalearningunitedvisual, maaz2024videochatgptdetailedvideounderstanding, dai2023instructblipgeneralpurposevisionlanguagemodels,zhang2023videollamainstructiontunedaudiovisuallanguage}.
Existing work primarily focuses on white-box attacks, where information about the target model is accessible. However, the transferability of adversarial video samples across V-MLLMs remains unexplored, which is a more common and practical setting in real-world scenarios. It is still uncertain whether the adversarial videos generated on the source model can effectively attack other target models, posing significant security risks to the deployment of V-MLLMs in real-world applications.



In this paper, we pioneer an investigation into the transferability of adversarial video samples across V-MLLMs. Through detailed analysis in Sec. \ref{sec:moti}, we think previous methods have these shortcomings: (1) lacking generalization in perturbing video features, (2) focusing only on sparse key-frames, and (3) failing to integrate multimodal information. FMM attack \cite{li2024fmmattackflowbasedmultimodaladversarial} is the first proposed white-box attack method targeting V-MLLMs. It utilizes flow-based temporal mask to select key-frames and applies perturbations to these frames. FMM attack performs well in the white-box setting but has limited transferability in the black-box setting.  
FMM attack heavily relies on the video features, which causes the generated perturbations to overfit to the video features extracted by the surrogate model, thereby reducing their generalizability.
Additionally, since FMM attack applies perturbations only to key-frames, it cannot ensure that all frames sampled by the target model are perturbed. 
Taking low-level image features into account can help with improving transferability  of adversarial samples.
Previous image-to-video cross-modal attacks \cite{wei2021crossmodaltransferableadversarialattacks, wang2023global, kim2023breakingtemporalconsistencygenerating} demonstrate the possibility of using image models as surrogates to attack video models in the black-box setting. 
However, these traditional attack methods typically focus on the video classification tasks with vision-only models, failing to integrate multimodal information.

To address these limitations, we propose a highly transferable attack method, named as Image To Video MLLM (I2V-MLLM) attack (see Fig. \ref{fig:1}). 
In I2V-MLLM, we utilize an image-based multimodal large language model (I-MLLM) as a surrogate model to craft adversarial video samples without accessing the internals of target V-MLLMs. Specifically, we extract key-frames from videos and send them into an I-MLLM to obtain adversarial perturbations. 
Multimodal interactions and spatiotemporal information are integrated to disrupt video representations within the latent space, improving adversarial transferability. Additionally, perturbation propagation technique is introduced to handle different unknown frame sampling strategies used by V-MLLMs. 

 We conduct various experiments on three well-established datasets, MSVD-QA \cite{xu2017video}, MSRVTT-QA \cite{xu2017video}, and ActivityNet-200 \cite{Heilbron_Escorcia_Ghanem_Niebles_2015} to evaluate the performance of our proposed I2V-MLLM attack in multiple video-text multimodal tasks. The experimental results demonstrate that our method can generate adversarial videos with strong transferability across different V-MLLMs (Chat-Univi \cite{jin2024chatuniviunifiedvisualrepresentation}, LLava-Next-Video \cite{zhang2024videoinstructiontuningsynthetic}, VideoChat, Video-LLaMA \cite{zhang2023videollamainstructiontunedaudiovisuallanguage}), and achieve competitive performance with white-box attacks against V-MLLMs. Our main contributions are summarized as follows: 
\begin{itemize}
\item We explore the transferable adversarial attack on four V-MLLMs and analyze the reasons for the low transferability when using existing methods to generate adversarial video samples (see Sec. \ref{sec:moti}). To the best of our knowledge, this is the first work to explore black-box attacks on V-MLLMs.
\item We propose a highly transferable attack method, named I2V-MLLM, for V-MLLMs using I-MLLMs to generate adversarial video samples (see Sec. \ref{sec:attack}). The adversarial videos generated by this method can effectively disrupt different V-MLLMs, significantly degrading their performance on multiple video-text multimodal tasks.
\item  We conduct extensive experiments on four different V-MLLMs using MSVD-QA, MSRVTT-QA, and ActivityNet-200 (see Sec. \ref{sec:exp_main} and Appendix B). The results demonstrate that our proposed attack method has strong transferability across V-MLLMs. 
\end{itemize}

\section{Related work}
\label{sec:formatting}

\subsection{Multimodal large language models }

MLLMs typically consist of a vision model, a pretrained LLM, and a projector that translates visual information into textual representations that the LLM can process. Currently, MLLMs can be categorized into image-based and video-based types. I-MLLMs  \cite{dai2023instructblipgeneralpurposevisionlanguagemodels, liu2023visualinstructiontuning, zhu2023minigpt4enhancingvisionlanguageunderstanding, alayrac2022flamingovisuallanguagemodel, awadalla2023openflamingoopensourceframeworktraining, hu2023blivasimplemultimodalllm, bai2023qwenvlversatilevisionlanguagemodel} are designed to handle image-text inputs. They excel in tasks such as visual question answering, image captioning, and more. V-MLLMs extend the capabilities of I-MLLMs by incorporating temporal modules that allow them to understand and process video input. This enables them to perform tasks like video question answering (VideoQA), spatiotemporal localization, and video captioning. For example, Chat-UniVi \cite{jin2024chatuniviunifiedvisualrepresentation} extracts specific frames from videos and utilizes DPC-KNN \cite{Du_Ding_Jia_2016} to group these frames into distinct events, Video-LLaMA \cite{zhang2023videollamainstructiontunedaudiovisuallanguage} employs sequential encoding to capture temporal relationships among video frames, VideoChatGPT \cite{maaz2024videochatgptdetailedvideounderstanding} applies temporal pooling to video features to extract temporal information. These methods equip the models with the capability to capture and interpret temporal dynamics, thus enabling a more comprehensive understanding of video content.

\subsection{Adversarial attacks on MLLMs}

Despite the impressive performance, MLLMs are highly susceptible to adversarial attacks \cite{zhao2023evaluatingadversarialrobustnesslarge, luo2024imageworth1000lies, cui2023robustnesslargemultimodalmodels, tu2023many, zhang2024avibench, bailey2023image, lu2023set}. 
For I-MLLMs, several studies have assessed their vulnerabilities to adversarial attacks. \cite{fu2023misusingtoolslargelanguage} introduces Trojan-like images that force the target models to invoke malicious external tools or APIs specified by the attacker. \cite{dong2023robustgooglesbardadversarial} utilizes open-source MLLMs to generate transferable adversarial examples capable of attacking closed-source commercial models like Bard~\cite{google2023gemini}, Bing Chat \cite{microsoft2024bingchat}, and GPT-4V \cite{openai2024gpt4v}, thereby showing high transferability of adversarial examples across MLLMs. \cite{gu2025improving} introduces DynVLA Attack, which applies dynamic perturbations to the vision-language connector to improve the generalization of adversarial attacks across different alignment strategies. While extensive studies have explored adversarial attacks on I-MLLMs, there has been little exploration in the domain of V-MLLMs. \cite{li2024fmmattackflowbasedmultimodaladversarial} proposes a flow-based adversarial attack strategy for white-box attacks on V-MLLMs. However, in real-world scenarios, the internal architectures and parameters of V-MLLMs are usually inaccessible to users. To address this, we focus on exploring methods for conducting adversarial attacks on V-MLLMs in a black-box setting.

\subsection{Adversarial attacks on video models}

Current video models have diverse applications, including autonomous vehicles, video verification, security, and other fields. However, these models remain vulnerable to adversarial attacks \cite{li2018adversarial, xie2022universal, jiang2019black, wei2020heuristic, cao2024logostylefoolvitiatingvideorecognition}. For example,  Universal 3D perturbations (U3D) \cite{xie2022universal} deceive video classifiers by generating a universal perturbation for all input videos, while StyleFool \cite{cao2023stylefool} introduces an unrestricted perturbation to attack video classification systems through style transfer. Recent studies also explore cross-modal attack methods from image models to video models \cite{wei2021crossmodaltransferableadversarialattacks, kim2023breakingtemporalconsistencygenerating, wang2023global}, yielding promising results. However, these attacks primarily target video classification tasks, which do not account for interactions between visual and textual modalities. In contrast, V-MLLMs integrate both visual and textual information, rendering these methods unsuitable for such models. To tackle this issue, our method introduces multimodal interactions in adversarial video crafting, aligning seamlessly with how V-MLLMs operate.

\section{Methodology}


\subsection{Preliminary}

Given a video sample \( V \in \mathcal{V} \)  with \( M \) associated QA pairs \( \{(q_m, a_m)\}_{m=1}^M \), where \( q_m \) is the \( m \)-th question and \( a_m \) is the corresponding answer. Let \( F \) denote the I-MLLM (e.g., BLIP-2 \cite{li2023blip2bootstrappinglanguageimagepretraining}, MiniGPT-4 \cite{zhu2023minigpt4enhancingvisionlanguageunderstanding}) and \( G \) denote the V-MLLM (e.g., Video-LLaMA \cite{zhang2023videollamainstructiontunedaudiovisuallanguage}, Chat-UniVi. We use \( G(V, q) \) to denote the answer generated by the V-MLLM for the given video \( V \) and question \( q \). The goal of our proposed attack is to generate an adversarial example \( V_{adv} = V + \delta' \) using \( F \), which can cause \( G \) to produce an answer \( G(V_{adv}, q_i) \) that differs significantly from the correct answer \( a_i \), without accessing the parameters or structure of \( G \), where \( \delta' \) denotes the adversarial perturbations specifically tailored for \( V \). To ensure that the adversarial perturbation \( \delta' \) is imperceptible, we restrict it by \( ||\delta'||_\infty \leq \epsilon \), where \( || \cdot ||_\infty \) denotes the \( L_\infty \) norm, and \( \epsilon \) is a constant for the norm constraint. We utilize the evaluation model \(E\) (i.e., GPT-4o-mini~\cite{openai2023gpt4omini}) to assess whether the generated answer aligns with the reference answer. We aim to find imperceptible adversarial perturbations that minimize the number of correct responses, formulated as follows:

\begin{equation}
\underset{\delta'}{\text{argmin}} \frac{1}{M}\sum_{i=1}^{M} E(G(V + \delta', q_i), a_i), \text{s.t.} \ \ ||\delta'||_\infty \leq \epsilon ,
\label{eq:0}
\end{equation}
where \( E(\cdot, \cdot) \) is the evaluation model’s judgment function, which outputs 1 if they match, and 0 otherwise.

\subsection{Motivation} \label{sec:moti}

To explore the transferability of adversarial videos across V-MLLMs, we first conduct an investigation of existing attack methods.
Based on the experimental results (in Tab. \ref{table:0}), we attribute their poor transferability to the following limitations:
(1) focusing only on sparse key-frames, (2) lacking generalization in perturbing video features, and (3) failing to integrate multimodal information. 

\textbf{Focusing only on sparse key-frames.} 
FMM attack exhibits limited performance when the key-frame ratio is low (see Appendix A). This is because FMM selects key-frames based on optical flow and only perturbs these frames, while V-MLLMs typically sample frames sequentially, making it difficult to ensure that all frames extracted by the target model are perturbed.
To address this, we first modify the FMM attack by replacing the sparse spatial perturbation with full perturbation on the key-frames sampled by V-MLLMs, which we call the Vanilla attack. 
While this adjustment improves white-box performance, the transferability still remains constrained. 
To further enhance transferability, we propagate perturbations from key-frames across the entire video, leading to improved transferability, as shown in rows 1, 2, 4, and 5 of Tab. \ref{table:0}. 

\textbf{Lacking generalization in perturbing video features.} 
Adversarial perturbations generated based on certain V-MLLM can overfit to specific video module, limiting their generalization to other V-MLLMs.
To improve transferability, we focus on lower-level image features. 
The I2V attack \cite{wei2021crossmodaltransferableadversarialattacks}, which perturbs each video frame to disrupt image features, demonstrates improved transferability when using image models as surrogates to craft adversarial video samples, as shown in rows 3, 4, and 5 of Tab. \ref{table:0}.

\textbf{Failing to integrate multimodal information.} 
The I2V attack shows a limited improvement in transferability, as it was originally designed for video classification and does not account for the multimodal interactions, which is essential for V-MLLMs. 
Therefore, we propose using an I-MLLM as a surrogate, integrating multimodal interaction information into the process of generating adversarial video samples, which leads to a significant improvement in transferability, as demonstrated in rows 3 and 6 of Tab. \ref{table:0}.

In summary, we propose using I-MLLMs as surrogates to generate adversarial videos that incorporate multimodal interactions. In addition, we introduce a perturbation propagation method to handle different unknown frame sampling strategies. 
The I2V-MLLM results in Tab. \ref{table:0} demonstrate the strong transferability of our method across different V-MLLMs. 
More discussions can be found in Appendix A. The following sections detail the I2V-MLLM attack.

\begin{table}[t]
\centering
\resizebox{\linewidth}{!}{
\renewcommand{\arraystretch}{1.2} 
\begin{tabular}{cccccc}
\hline
\multirow{2}{*}{\textbf{Attack}} & \multicolumn{4}{c}{\textbf{Target Model}} & \multirow{2}{*}{\textbf{AASR}}\\ \cline{2-5}
& \textbf{Chat-UniVi} & \textbf{LLaVA-NeXT-Video}& \textbf{VideoChat} & \textbf{Video-LLaMA}  &\\ \hline
\textbf{FMM} & 8.70& 18.76& 13.84& 27.93* &17.31\\
\textbf{Vanilla} & 11.62& 25.31& 15.10& 64.14* &29.04\\
\textbf{I2V} & 32.17& 33.39& 41.13& 36.51&35.80\\ \hline
\textbf{FMM w/ Prop.} & 14.54 & 25.31& 14.62& 27.93* &20.60\\
\textbf{Vanilla w/ Prop.} & 14.94& 32.24& 17.50& 64.14*&32.21\\
\rowcolor{gray!20} \textbf{I2V-MLLM}& \textbf{48.39}& \textbf{45.54}& \textbf{63.09}& \textbf{74.91}&\textbf{57.98}\\ \hline
\end{tabular}
}
\caption{
Attack success rates (ASR, \%)  on the MSVD-QA validation set for Zero-Shot VideoQA tasks. \textbf{FMM} and \textbf{I2V} denote attack methods from \cite{li2024fmmattackflowbasedmultimodaladversarial} and \cite{wei2021crossmodaltransferableadversarialattacks}, respectively. \textbf{Vanilla} attack applies full perturbations on all key-frames sampled by V-MLLMs. \textbf{Prop.} denotes perturbation propagation. * indicates white-box attacks. AASR represents the average ASR across all target models for each surrogate model. A higher AASR indicates better adversarial transferability.
}
\label{table:0}
\vspace{-0.5cm}
\end{table}

\subsection{I2V-MLLM Attack} \label{sec:attack}

The proposed I2V-MLLM attack utilizes an I-MLLM to produce adversarial video samples, targeting image-to-video cross-modal black-box attacks on V-MLLMs with significant transferability. By manipulating the intermediate features of vision models and projectors of I-MLLMs, our approach generates adversarial video samples that interfere with the intermediate features of black-box V-MLLMs.  The I2V-MLLM attack algorithm is illustrated in Appendix C, consists of three components: vision model attack, projector attack, and perturbation propagation.


\subsubsection{Vision Model Attack} \label{sec:vision_model_attack}

To enhance generalization in perturbing video features, I2V-MLLM disrupts both image features and spatiotemporal information extracted by the vision model. We first split the video \( V \) into \( K \) clips: \( V = \{v^1, v^2, \ldots, v^K\} \), where
\( K = \text{total number of frames} \times \text{key-frame ratio} \, \beta \). We select the first frame \( x^k \) from each clip \( v^k \) as the key-frame, resulting in \( K \) key-frames, \( X = \{x^1, x^2, \ldots, x^K\} \), each capturing the essential information of their respective clips. We extract spatiotemporal embeddings of \(X\) using the vision model. This model independently encodes the \( K \) frames, producing frame-level embeddings \(F_V(X) \in \mathbb{R}^{K \times N \times D_1} \), where \(F_V(\cdot) \) denotes the encoder of the vision model, \( N \) is the number of patches per frame, and \( D_1 \) is the dimension of the embeddings. Frame-level embeddings are average-pooled along the temporal dimension to obtain  temporal embeddings \(F_V^t(X) \in \mathbb{R}^{N \times D_1} \), which implicitly incorporates temporal information of $K$ frames. 
Similarly, the frame-level embeddings are average-pooled along the spatial dimension to obtain  spatial embeddings \(F_V^s(X) \in \mathbb{R}^{K \times D_1} \), which incorporate the spatial information of $K$ frames. 
The temporal and spatial embeddings are concatenated to obtain the original spatiotemporal embeddings\(
F_V^{ts}(X) = [F_V^t(X), F_V^s(X)] \in \mathbb{R}^{(N + K) \times D_1}.
\)
For the adversarial input \( X_{adv} = \{x^1 + \delta^1, x^2 + \delta^2, \ldots, x^K + \delta^K\} \), we can similarly obtain the adversarial spatiotemporal embeddings \(F_V^{ts}(X_{adv}) \). 
To disrupt the spatiotemporal features, I2V-MLLM optimizes the adversarial perturbations by minimizing the cosine similarity between the original and the adversarial spatiotemporal embeddings: 
\begin{equation}
\mathcal{L}_{V} = \sum_{i=1}^{N + K}\frac{ Cos(F_V^{ts}(X)_{i},F_V^{ts}(X_{adv})_{i})}{N + K} ,
\label{eq:1}
\end{equation}
where \(F_V^{ts}(X)_i \) and \(F_V^{ts}(X_{adv})_i \) represent the \( i \)-th elements in the spatiotemporal embeddings of the original and the adversarial video frames, respectively. 

\subsubsection{Projector Attack}

To further disrupt V-MLLMs' capacity for video-text multimodal tasks, I2V-MLLM interferes with the intermediate feature of the projector (e.g. Q-Former \cite{li2023blip2bootstrappinglanguageimagepretraining}), which plays an essential role in aligning visual and textual representations. We feed the projector with the original frame-level embeddings \(F_V(X) \), the adversarial frame-level embeddings \(F_V(X_{adv}) \) from the vision model, and the caption set \( T = \{t_1, t_2, \ldots, t_M\} \). After multimodal alignment, they are transformed into the original visual features \( F_P^v(X) \in \mathbb{R}^{N_1 \times D_2} \), the adversarial visual features \( F_P^v(X_{adv}) \in \mathbb{R}^{N_1 \times D_2} \), and the textual features \( F_P^t(T) \in \mathbb{R}^{N_2 \times D_2} \). Here, \( N_1 \) and $N_2$ represent the number of visual features and the textual features, respectively. And \( D_2 \) denotes the dimension of these features. The captions are complete sentences generated based on the question \( q \) and the answer \( a \) using GPT-4o-mini \cite{openai2023gpt4omini}. For example, given the question \( q \): `What is the man doing?' and the answer \( a \): `eat', the corresponding caption \(t\) would be: `The man is eating.'
To perturb the image features aligned with the text, I2V-MLLM optimizes the adversarial perturbations by minimizing the cosine similarity between the original and the adversarial visual features:
\begin{equation}
\mathcal{L}_{P_{v}} = \sum_{n_1=1}^{N_1}  \frac{Cos(F_P^v(X)_{n_1}, F_P^v(X_{adv})_{n_1})}{N_1},
\label{eq:2}
\end{equation}
where \( F_P^v(X)_{n_1} \) and \(F_P^v(X_{adv})_{n_1} \) are the \( n_1 \)-th visual feature of the original and the adversarial video frames, respectively. 
To disrupt multimodal interactions between adversarial frames and text, I2V-MLLM optimizes the adversarial perturbations by minimizing the cosine similarity between the adversarial visual features and the textual features: 
\begin{equation}
\mathcal{L}_{P_{v2t}} = \sum_{n_1=1}^{N_1}\sum_{n_2=1}^{N_2}\frac{Cos(F_P^v(X_{adv})_{n_1}, F_P^t(T)_{n_2})}{N_1N_2} ,
\label{eq:3}
\end{equation}
where \(F_P^t(T)_{n_2}\) is the \(n_2\)-th textual feature of \(T\).  The total loss function for projector  is 
\begin{equation}
\mathcal{L}_{P} =  \mathcal{L}_{P_{v}} +  \mathcal{L}_{P_{v2t}}.
\label{eq:4}    
\end{equation}

\begin{table*}[t]
\centering
\renewcommand{\arraystretch}{1.2} 
\resizebox{\linewidth}{!}{
\begin{tabular}{ccllllllllllllc}
\hline
\multirow{2}{*}{\textbf{Attack}} & \multirow{2}{*}{\textbf{Surrogate Model}} & \multicolumn{3}{c}{\textbf{Chat-UniVi}} & \multicolumn{3}{c}{\textbf{LLaVA-NeXT-Video}} & \multicolumn{3}{c}{\textbf{VideoChat}} & \multicolumn{3}{c}{\textbf{Video-LLaMA}} & \multirow{2}{*}{\textbf{AASR}}\\ \cline{3-14}
                        &                                  & ASR $\uparrow$& Acc.  $\downarrow$& Score $\downarrow$& ASR $\uparrow$& Acc.  $\downarrow$& Score $\downarrow$& ASR $\uparrow$& Acc.  $\downarrow$&Score $\downarrow$& ASR $\uparrow$& Acc.  $\downarrow$&Score $\downarrow$ &\\ \hline
\textbf{Clean}                 & /                                & /     & 60.89    & 3.34      & /     & 48.95    & 2.90      & /     & 60.24    & 3.42      & /     & 53.81    & 3.09       & / \\ \hline
\multirow{4}{*}{\textbf{FMM}}      & Chat-UniVi                     & 16.00* & 57.41* & 3.18* & 16.33& 50.38& 2.93& 13.21& 60.75& 3.39& 21.47& 53.31& 3.06 &16.76\\ 
                        & LLaVA-NeXT-Video                 & 9.22& 60.65& 3.34& 20.48* & 47.84* & 2.83* & 13.49& 60.30& 3.38& 21.32& 53.43& 3.05 &16.13\\ 
                        & VideoChat                       & 8.12& 61.81& 3.38& 15.38& 51.30& 2.98& 14.62* & 59.91* & 3.35* & 20.74& 54.08& 3.09 &14.72\\ 
                        & Video-LLaMA                      & 8.70& 61.53& 3.36& 18.76& 49.40& 2.89& 13.84& 60.20& 3.38& 27.93* & 48.39* & 2.84* &17.31\\ \hline
\multirow{4}{*}{\textbf{Vanilla}}& Chat-UniVi                     & 56.34*& 30.40*& 1.88*& 12.64& 48.60& 2.87& 6.00& 59.05& 3.35& 21.78& 53.53& 3.06&24.19\\ 
                        & LLaVA-NeXT-Video                 & 9.33& 59.05& 3.30& 52.45*& 24.96*&1.53*& 5.15& 59.97    & 3.37      & 24.94& 50.82& 2.95&22.97\\ 
                        & VideoChat                       & 7.35& 59.61& 3.38& 7.93& 49.90& 2.91& 68.90*& 23.81*& 1.67*& 20.74& 54.15& 3.09&26.23\\ 
                        & Video-LLaMA                      & 11.62& 58.88    & 3.26      & 25.31& 41.01    & 2.52      & 15.10& 59.33    & 3.34      & 64.14*& 23.88* & 1.72* &29.04\\ \hline
\textbf{I2V}                    & CLIP-L/14                        & 32.17& 51.53& 2.92& 33.39& 43.63& 2.60& 41.13& 49.57& 2.91& 36.51& 46.49& 2.71&35.80\\ \hline
 \rowcolor{gray!20} & BLIP-2                          & \textbf{48.39}& \textbf{34.72}& \textbf{2.03}& 45.54& 29.33& 1.94& \textbf{63.09} & \textbf{26.08} & \textbf{1.82} & \textbf{74.91}& \textbf{17.07}& \textbf{1.39} &\textbf{57.98}\\ 
         \rowcolor{gray!20}\textbf{I2V-MLLM}& InstructBLIP                     & 45.74& 35.10& 2.16& 44.61& 30.64& 2.13& 54.26 & 31.99 & 2.10 & 69.90 & 20.58 & 1.58 &53.63\\ \rowcolor{gray!20}
               & MiniGPT-4                        & 43.58& 37.02& 2.21& \textbf{46.50}& \textbf{27.98}& \textbf{1.76}& 56.51 & 30.49 & 2.06 & 68.92 & 21.37 & 1.60 & 53.88\\ \hline
\end{tabular}
}
\caption{The results on the \textbf{MSVD-QA} for Zero-Shot VideoQA tasks. ASR (\%) indicates attack success rate. Acc.(\%) denotes the accuracy of the model's predictions, while the Score represents GPT Score, which assesses the model and assigns a relative score to the predictions on a scale of 0 to 5. AASR represents the average ASR across all target models for each surrogate model.  A higher AASR indicates better adversarial transferability. The best ASR for each target model under \textbf{black-box attacks} is highlighted in \textbf{bold}. * indicates white-box attacks for reference.}

\label{table:2}
\vspace{-0.25cm}
\end{table*}

\begin{table*}[t]
\centering

\renewcommand{\arraystretch}{1.2} 
\resizebox{\linewidth}{!}{
\begin{tabular}{ccllllllllllllc}
\hline
\multirow{2}{*}{\textbf{Attack}} & \multirow{2}{*}{\textbf{Surrogate Model}} & \multicolumn{3}{c}{\textbf{Chat-UniVi}} & \multicolumn{3}{c}{\textbf{LLaVA-NeXT-Video}} & \multicolumn{3}{c}{\textbf{VideoChat}} & \multicolumn{3}{c}{\textbf{Video-LLaMA}} & \multirow{2}{*}{\textbf{AASR}}
\\ \cline{3-14}
                        &                                  & ASR $\uparrow$& Acc.  $\downarrow$& Score $\downarrow$& ASR $\uparrow$& Acc.  $\downarrow$& Score $\downarrow$& ASR $\uparrow$& Acc.  $\downarrow$& Score $\downarrow$& ASR $\uparrow$& Acc.  $\downarrow$& Score $\downarrow$ &\\ \hline
\textbf{Clean}                 & /                                & /     & 39.62& 2.51& /     & 29.17& 2.06& /     & 38.92& 2.50& /     & 31.42& 2.17 & /     \\ \hline
\multirow{4}{*}{\textbf{FMM}}      & Chat-UniVi                     & 23.39*& 36.85*& 2.36*& 24.79& 31.60& 2.17& 9.04& 39.44& 2.53& 32.50& 32.03& 2.20 &22.43\\ 
                        & LLaVA-NeXT-Video                 & 13.20& 40.01& 2.52& 28.62*& 29.90*& 2.09*& 8.52& 39.24& 2.51& 32.27& 31.94& 2.19 &20.65\\ 
                        & VideoChat                       & 12.83& 40.52& 2.54& 25.29& 31.10& 2.15& 15.15*& 37.99*& 2.46*& 30.48& 32.56& 2.21 &20.94\\ 
                        & Video-LLaMA                      & 12.72& 40.80& 2.55& 27.92& 30.25& 2.12& 8.16& 39.71& 2.53& 37.38*& 29.60*& 2.07*&21.55\\ \hline
\multirow{4}{*}{\textbf{Vanilla}}& Chat-UniVi                     & 52.19*& 23.10*& 1.68*& 25.85& 30.09& 2.10& 9.58& 39.52& 2.52& 32.32& 32.24& 2.21 &29.99\\ 
                        & LLaVA-NeXT-Video                 & 13.07& 41.08& 2.56& 65.90*& 15.06*& 1.46*& 8.50& 39.56& 2.53& 35.23& 30.41& 2.13 &30.68\\ 
                        & VideoChat                       & 12.64& 41.35& 2.57& 25.27& 30.72& 2.14& 63.46*& 16.69*& 1.42*& 30.84& 32.58& 2.21 &33.05\\ 
                        & Video-LLaMA                      & 14.33& 40.78& 2.55& 37.91& 26.14& 1.91& 8.83& 39.53& 2.52& 64.11*& 18.07*& 1.53*&31.29\\ \hline
\textbf{I2V}                    & CLIP-L/14                        &       30.05&          34.53&           2.28&       35.62&          26.96&           1.96&       18.83&          38.59&           2.50&       36.16&          30.16&           2.12 &30.17\\ \hline
\rowcolor{gray!20}  &  BLIP-2                          & \textbf{47.93}& \textbf{28.42}& \textbf{2.00}& \textbf{53.78}& \textbf{19.34}& \textbf{1.58}& \textbf{62.38}& \textbf{18.72}& \textbf{1.57}& \textbf{78.95}& \textbf{10.68}& \textbf{1.17} & \textbf{60.76}\\ 
                   \rowcolor{gray!20} \textbf{I2V-MLLM }   & InstructBLIP                     & 45.37& 31.88& 2.14& 50.96& 21.72& 1.70& 54.78& 22.66& 1.76& 73.04& 13.52& 1.34 &56.04\\ 
                  \rowcolor{gray!20}   & MiniGPT-4                        & 46.60& 30.94& 2.11& 49.47& 21.12& 1.67& 56.41& 21.95& 1.73& 73.28& 13.63& 1.32 &56.44\\ \hline
\end{tabular}
}
\caption{The results on the \textbf{MSRVTT-QA}. The corresponding metrics and settings are consistent with those in Tab. \ref{table:2}.}
\label{table:3}
\vspace{-0.5cm}
\end{table*}

\subsubsection{Optimization and Perturbation Propagation}
To maximize the efficacy of the adversarial attack, we combine the losses \( \mathcal{L}_V \) and \(\mathcal{L}_P\) into a unified objective. This combined loss ensures that both the vision model and the projector's intermediate features are significantly perturbed. The unified loss is formulated as: 
\begin{equation}
\mathcal{L}_{total} = \lambda_1\mathcal{L}_V + \lambda_2\mathcal{L}_P ,
\label{eq:5}    
\end{equation}
where $\lambda_1$ and $\lambda_2$ correspond to the two loss weights, which aim to balance them during the optimization.

We optimize \( \delta_k \) according to the following expression:
\begin{equation}
\delta^k = \underset{\delta^k}{\arg\min} (\mathcal{L}_{total}), s.t. ||\delta^k||_\infty \leq \epsilon, k = 1, \ldots, K.
\label{eq:6}    
\end{equation}

Finally, we replicate \(\delta^k \) to match the length of  its corresponding video clip \(v^k \), obtaining perturbed clip  \(\delta'^k\).  The adversarial video is then constructed by applying pixel-wise addition to combine these perturbed clips with the original ones: \(
V_{adv} = V + \delta' = \{v^1 + \delta'^1, v^2 + \delta'^2, \ldots, v^K + \delta'^K\}
\). We term  this approach Direct Propagation (DP). Due to the high similarity between consecutive frames, DP proves to be a simple yet effective method (see Sec. 4.3).  

\section{Experiment}

\subsection{Experimental setting}

In this section, we present the experimental setting, including datasets, models, attack setting and metrics.

\textbf{Datasets and models.} Referring to the quantitative benchmarking framework proposed in  \cite{maaz2024videochatgptdetailedvideounderstanding}, we evaluate our I2V-MLLM attack on Zero-Shot VideoQA tasks using the validation set of MSVD-QA and MSRVTT-QA. We perform the proposed method on three I-MLLMs: BLIP-2, InstructBLIP \cite{dai2023instructblipgeneralpurposevisionlanguagemodels}, and MiniGPT-4. Our method is evaluated on four different V-MLLMs: Chat-UniVi, LLaVA-NeXT-Video, VideoChat, and Video-LLaMA \cite{zhang2023videollamainstructiontunedaudiovisuallanguage}, each with a Vicuna-7B \cite{vicuna2023} as the LLM.

\textbf{Attack setting.} In I2V-MLLM, we employ the projected gradient descent (PGD) \cite{madry2019deeplearningmodelsresistant} with a perturbation bound of $\epsilon = 16$, an iteration number of $I = 50$, and a step size of $\alpha = 1$ for the attack process. The parameters $\lambda_1$ and $\lambda_2$ are both set to $1$, and the key-frame ratio $\beta$ is set to $30\%$. I2V attack, utilizing CLIP-L/14 \cite{radford2021learningtransferablevisualmodels} as the surrogate model, applies tailored perturbations to each frame of the video.
For a fair comparison, the PGD parameters ($\epsilon = 16$, $I = 50$ and $\alpha = 1$) in FMM, Vanilla, and I2V attacks maintain the same for our method. 
Additionally, in the FMM setup, the key-frame ratio $\beta$ is also set to $30\%$. 
All the experiments are conducted on an NVIDIA-A6000 GPU. 

\textbf{Metrics.} 
We use Attack Success Rate (ASR) to evaluate the effectiveness of adversarial examples on Zero-Shot VideoQA tasks.
It measures the percentage of successful attacks on questions the model answered correctly for clean videos. 
Answer correctness is evaluated using GPT-4o-mini, which checks whether the model's prediction semantically aligns with the ground truth.
We also provide the average ASR (AASR) across all evaluated V-MLLMs. A higher ASR or AASR indicates better adversarial transferability. 
To evaluate the model's overall performance when encountering adversarial videos, we further employ GPT-assisted methods \cite{maaz2024videochatgptdetailedvideounderstanding} to assess Accuracy (Acc.) and GPT-Score. Specifically, accuracy (Acc.) refers to the model’s prediction accuracy, while the GPT score (Score) assesses the quality of the model’s predictions, assigning a relative score on a scale from 0 to 5. GPT-4o-mini is used for evaluation due to its strong text understanding and cost efficiency.
For detailed explanations of the metrics, see Appendix B.1.

\subsection{Attack performance} \label{sec:exp_main}
In this section, we compare our proposed I2V-MLLM attack with the FMM, Vanilla, and I2V attacks. The results, summarized in Tab. \ref{table:2} and Tab. \ref{table:3}, provide a quantitative comparison of the ASR, AASR, Acc., and GPT Score for the MSVD-QA and MSRVTT-QA datasets, respectively.

\textbf{Evaluation of ASR.} 
As shown in Tab. \ref{table:2} and Tab. \ref{table:3}, I2V-MLLM achieves the highest AASR compared to previous attack methods, achieving AASR of 57.98\%, 53.63\%, and 53.88\% for MSVD-QA and 60.76\%, 56.04\%, and 56.44\% for MSRVTT-QA when taking BLIP-2, InstructBLIP, MiniGPT-4 as surrogate models, respectively, significantly outperforming previous attack methods. 
I2V-MLLM (BLIP-2) achieves the best attack performance on Video-LLaMA and near-best attack performance on other target models. It achieves ASRs of 48.39\%, 45.54\%, 63.09\%, and 74.91\% on MSVD-QA, and 47.93\%, 53.78\%, 62.38\%, and 78.95\% on MSRVTT-QA, respectively, outperforming both the FMM and I2V attacks while achieving performance comparable to the white-box Vanilla attack. 

\textbf{Evaluation of the quality of generated answers.} 
We also incorporate Acc. and GPT Score as metrics to better analyze the impact of adversarial videos on V-MLLM performance.
As shown in Tab. \ref{table:2} and Tab. \ref{table:3}, the proposed I2V-MLLM significantly reduces both Acc. and Scores across all target models, particularly for VideoChat and Video-LLaMA. 
On the MSVD-QA dataset, Acc. drops to 26.08\% and 17.07\%, while Scores fall to 1.82 and 1.39. 
On the MSRVTT-QA dataset, Acc. further declines to 18.72\% and 10.68\%, with Scores of 1.57 and 1.17, respectively.
Significant effects are also observed on Chat-UniVi and LLaVA-NeXT-Video.
These significant performance degradations highlight the destructive power of the I2V-MLLM attack, demonstrating its transferability and effectiveness across multiple V-MLLMs, while revealing the adversarial vulnerability of existing models, even in black-box settings.



\subsection{Ablation study}
In this section, we provide ablation studies on the loss functions, key-frame ratio \(\beta\), perturbation propagation and different projectors in I2V-MLLM attack. Experiments are conducted on the MSVD-QA for Zero-Shot VideoQA tasks. 

\begin{figure}[t]
    \centering
    \includegraphics[width=0.8\linewidth]{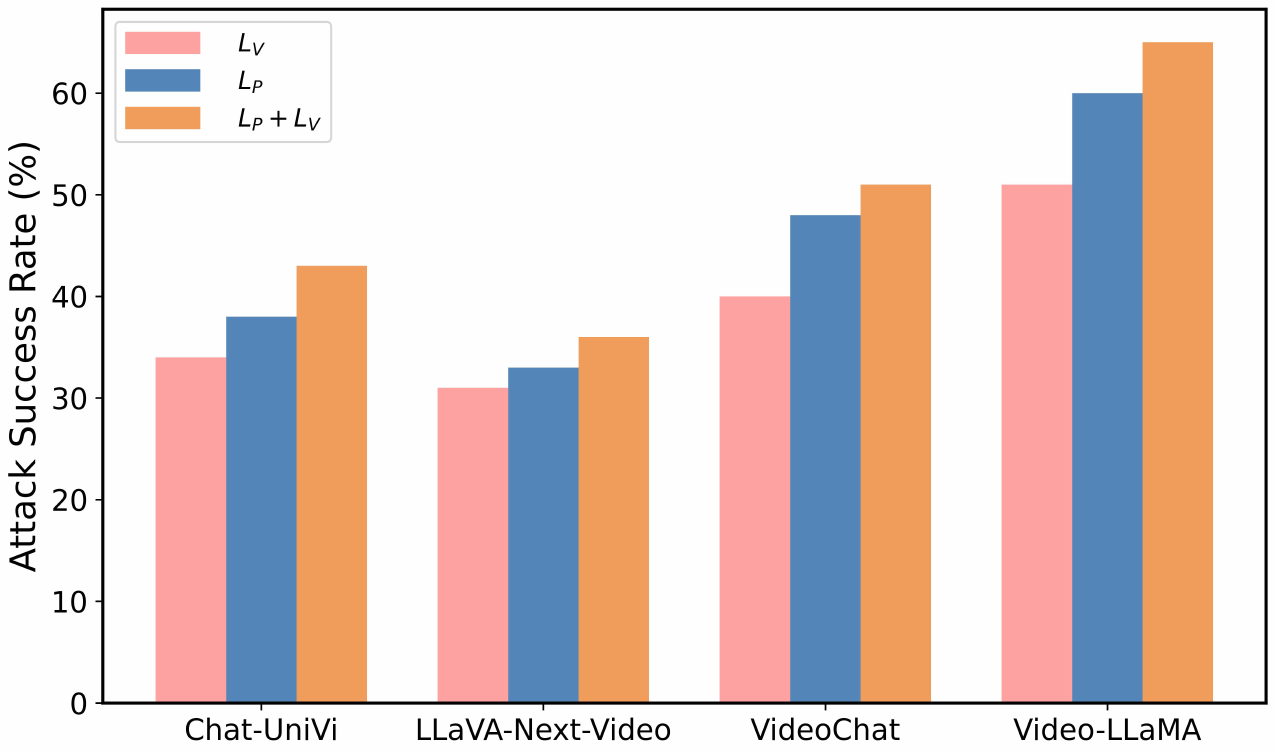}
    \caption{Attack success rates (ASR, \%)  of the I2V-MLLM attack with different loss functions.}
    \label{fig:3}
\vspace{-0.25cm}
\end{figure}

\textbf{Influence of loss functions.} 
In Fig. \ref{fig:3}, we provide ablation study on the components of the loss functions used in our I2V-MLLM.
The surrogate model is BLIP-2, and the generated adversarial videos are evaluated across four V-MLLMs.
It can be observed that using either \(\mathcal{L}_V\) or \(\mathcal{L}_P\) alone achieves satisfactory attack performance. 
Combining both, which simultaneously disrupts low-level image features and the alignment between visual and textual modalities, further enhances the attack performance.

\begin{figure}[t]
    \centering
    \includegraphics[width=0.8\linewidth]{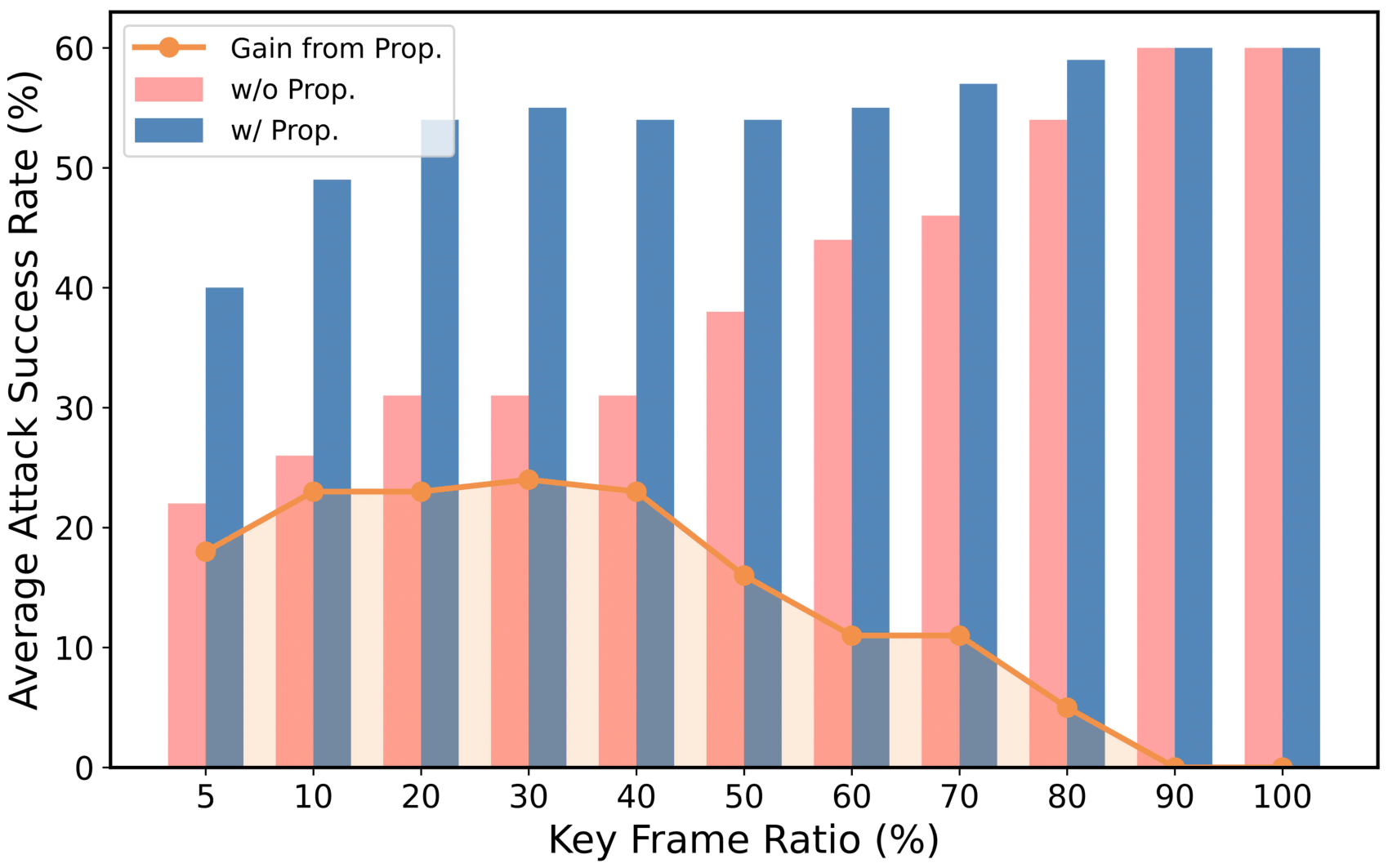}
    \caption{AASR (\%) of the I2V-MLLM attack with various key-frame ratios, comparing scenarios with and without perturbation propagation. `Prop.' represents `Propagation'.}
    \label{fig:4}
\vspace{-0.5cm}
\end{figure}

\textbf{Influence of key-frame ratio and propagation method.}
The number of key-frames used to optimize the perturbation, as well as the decision to propagate these perturbations across the entire video, significantly affects the transferability of adversarial video samples. 
Fig. \ref{fig:4} illustrates the results obtained with various key-frame ratios, comparing scenarios with and without perturbation propagation. 
It can be observed that as the key frame ratio increases, the generated adversarial samples show improved transferability.
Perturbation propagation substantially improves AASR by ensuring that all the frames extracted by unseen V-MLLMs are perturbed.
As illustrated by the gain curve in the Fig. \ref{fig:4}, the improvement from perturbation propagation initially rises with the key-frame ratio but then diminishes, reaching its maximum at 30\%. With an AASR already high at a 30\% key-frame ratio, further increases yield minimal gains, and perturbation propagation reaches its maximal benefit at this point. 
Therefore, we adopt a key-frame ratio of $\beta = 30\%$. 

\begin{table}[t]
\centering
\renewcommand{\arraystretch}{1.2} 
\resizebox{\linewidth}{!}{ 
\begin{tabular}{ccccc}
\hline
\textbf{Propagation Method}&\textbf{/}& \textbf{OFP}& \textbf{BP}& \textbf{DP}\\ \hline
Chat-UniVi&37.69& 34.03
& 44.63
& \textbf{48.39}
\\ 
LLaVA-NEXT-Video&23.34& 32.07
& 39.72
& \textbf{45.54}
\\
 VideoChat&23.67& 42.26
& 54.60& \textbf{63.09}
\\ 
 Video-LLaMA&45.89& 62.40& 71.23
& 7\textbf{4.91}
\\ \hline
 AASR& 32.65& 42.69
& 52.55
&\textbf{57.98}\\ \hline
\end{tabular}
}
\caption{Attack success rates (ASR, \%) of the I2V-MLLM attack with different perturbation propagation method.  ‘/’ represents no perturbation propagation.   A higher AASR indicates better adversarial transferability.}
\label{table:propagation_method}
\vspace{-0.5cm}
\end{table}

\begin{table}[t]
\centering
\renewcommand{\arraystretch}{1.2} 
\resizebox{\linewidth}{!}{ 
\begin{tabular}{ccccccc|c}
\hline
\textbf{ Attack}&\textbf{Surrogate}&  \textbf{Projector}&\textbf{C-U}& \textbf{L-N-V}& \textbf{V-C}& \textbf{V-L} &\textbf{AASR}\\ \hline
 \multirow{4}{*}{\textbf{Vanilla}}& C-U& /& 56.34*& 12.64& 6.00& 21.78&24.19\\
 & L-N-V& /& 9.33& 52.45*& 5.15& 24.94&22.97\\
 & V-C& /& 7.35& 7.93& 68.90*& 20.74&26.23\\
 & V-L & /& 11.62& 25.31& 15.10& 64.14*&29.04\\ \hline
   &LLaVA&  FC-Linear&41.31& 43.32& 50.31& 56.72&47.92\\ 
 &mPLUG-Owl&  FC-Linear&44.97& 42.07& 49.77& 57.16&46.74\\ 
 
 \textbf{I2V-MLLM}& BLIP-2&  Q-Former&\textbf{48.39}& 45.54& \textbf{63.09}& \textbf{74.91}&\textbf{57.98}\\ 
 & InstructBLIP&  Q-Former&45.74& 44.61& 54.26& 69.90&53.63\\
  & MiniGPT-4&  Q-Former&43.58& \textbf{46.50}& 56.51& 68.92&53.88\\ \hline
\end{tabular}
}
\caption{Ablation Study on I-MLLMs with different projectors. Attack success rates (ASR, \%)  on the MSVD-QA validation set for Zero-Shot VideoQA tasks. * indicates a white-box attack. A higher AASR indicates better adversarial transferability.  The highest attack performance for each target model in I2V-MLLM is shown in \textbf{bold}. \textbf{Note:} C-U: Chat-UniVi, L-N-V: LLaVA-NeXT-Video, V-C: VideoChat, V-L: Video-LLaMA.
  }
\label{table:projector}
\vspace{-0.25cm}
\end{table}

Different perturbation propagation methods may yield varying results. We set the key-frame ratio \(\beta = 
 30\%\) and test three distinct perturbation propagation methods: Direct Propagation (DP), Optical Flow-based  Propagation \cite{dosovitskiy2015flownet} (OFP), and Bidirectional Linear Interpolation Propagation \cite{dai2017deformable} (BP). The details of propagation methods are in Appendix B.1. As shown in Tab. \ref{table:propagation_method}, DP achieves the most significant improvement in AASR. Due to the high similarity between consecutive frames, DP proves to be a simple yet effective method. OFP suffers from added complexity and may distort the perturbation, resulting in lower effectiveness. BP's slightly lower performance stems from its reliance on interpolation, which may dilute perturbation intensity compared to DP's direct application. Therefore, we adopt DP in I2V-MLLM. 

\textbf{Influence of different projectors.}  \label{exp_projector}
The projectors of I-MLLMs are typically either FC-Linear (LLaVA \cite{liu2023visualinstructiontuning}, mPLUG-Owl \cite{ye2023mplug}) or Q-Former (BLIP-2, InstructBLIP, MiniGPT-4). FC-Linear maps the visual features extracted by the vision encoder into the latent space of the LLM, whereas Q-Former further aligns visual and textual features before passing them to the LLM, enabling richer multimodal interactions. When Q-Former-based I-MLLM is employed as the surrogate, the projector attack module's loss function is \( \mathcal{L}_{P} =  \mathcal{L}_{P_{v}} + \mathcal{L}_{P_{v2t}}\), where as for FC-Linear-based I-MLLM, it becomes \(\mathcal{L}_{P} =  \mathcal{L}_{P_{v}}\). As shown in Tab. \ref{table:projector}, irrespective of the choice of surrogate model and projector, the adversarial transferability of I2V-MLLM significantly surpasses that of the vanilla attack. I-MLLMs equipped with Q-Former consistently outperform those with FC-Linear in generating transferable adversarial samples. This underscore the critical role of multimodal interactions in adversarial attacks on V-MLLMs. Therefore, in our main experiments, we exclusively employ Q-Former-based I-MLLMs as surrogate models to fully demonstrate the effectiveness of the I2V-MLLM approach. 

Extended discussions on the \textbf{influence of} \textbf{step size}, \textbf{iteration number}, \textbf{weights of loss functions} and other experiments are detailed in Appendix B.2.



\section{Conclusion}
In this paper, we are the first to explore black-box transferable attacks on V-MLLMs. We conduct a thorough investigation of the limitations of existing methods, revealing that they exhibit lower transferability despite their impressive performance in white-box settings. Our findings underscore the need for specially designed transferable attacks tailored to V-MLLMs. We propose the I2V-MLLM attack, a highly transferable cross-modal attack that leverages the intermediate features of I-MLLMs and perturbation propagation to enhance the transferability of attacks targeting V-MLLMs. We hope our work will inspire further research aimed at evaluating and improving the robustness of V-MLLMs.

\section*{Acknowledgments}

Feng Zheng was supported by the National Natural Science Foundation of China (K23271004).
Xi Xiao and Linhao Huang were supported by the Natural Science Foundation of Guangdong Province (2025A1515011946) and the Major Key Project of PCL (PCL2023A06-4).
XJ and BH were supported by the NSFC General Program (Project No. 62376235) and the Guangdong Basic and Applied Basic Research Foundation (Project Nos. 2022A1515011652 and 2024A151501239).
\bibliography{aaai2026}
\clearpage
\maketitleappendix
\appendix
\setcounter{section}{0}
\input{supple}
\end{document}

%% file: supple.tex
\section{Motivation} \label{suppl_motivation}

To improve the transferability of attacks on V-MLLMs, we first conduct a thorough investigation into the shortcomings of existing methods, as showed in Tab. 1 in the main content. Based on the experimental results, we summarize the shortcomings of these existing methods as follows: (1) focusing only on sparse key-frames, (2) lacking generalization in perturbing video features, and (3) failing to integrate multimodal information. 

\textbf{Focusing only on sparse key-frames.} 
FMM attack exhibits limited performance when the key-frame ratio is low. This is because FMM selects key-frames based on optical flow and only perturbs these frames, while V-MLLMs typically sample frames sequentially, making it difficult to ensure that all frames extracted by the target model are perturbed.

\label{fmm_abl}
To validate the impact of the mismatch between FMM's key-frame selection and V-MLLMs' sampling strategies on the attack performance, we conduct an ablation study on the key-frame ratio in FMM attack. We perform a white-box FMM attack on Video-LLaMA. As demonstrated in Tab. \ref{table:video_fmm_k}, as the key-frame ratio in FMM increases, the number of adversarial frames sampled by Video-LLaMA also increases, leading to a corresponding improvement in ASR. This demonstrates that the method of selecting key-frames using optical flow in FMM is mismatched with the key-frame selection approach of V-MLLMs, and confirms that injecting perturbations into all frames that V-MLLMs will sample maximizes the attack effectiveness. 

\begin{table}[h]
\centering
\renewcommand{\arraystretch}{1.2} 
\resizebox{\linewidth}{!}{ 
\begin{tabular}{ccccccc}
\hline
Key frame ratio& 10& 30& 50& 70& 90&100\\ \hline
Sampled frames& 1.35& 2.67& 4.02& 5.87& 7.36&\textbf{8.00}\\ 
ASR& 23.12& 28.45& 34.31& 41.02 & 58.37&\textbf{64.14}\\ \hline
\end{tabular}
}
\caption{The ablation study on the key-frame ratio (\%) in the FMM attack is conducted to evaluate its impact on the attack performance. \textbf{Sampled frames} refer to the average number of adversarial frames that Video-LLaMA selects during its forward inference process for each video in the MSVD-QA validation set. (with a maximum of 8 frames).}
\label{table:video_fmm_k}
\end{table}

We replaced the flow-based mask with direct perturbations on all frames sampled by V-MLLMs, calling this the \textit{Vanilla} attack. The comparison between the FMM and Vanilla attacks in Tab. 1 in the main content shows a marked improvement in white-box performance after the adjustment. However, the transferability remains limited due to diverse frame-sampling strategies in V-MLLMs. Extending key-frame perturbations to the entire video further improves transferability, as shown in rows 1, 2, 4, and 5 of Tab. 1 in the main content. These results emphasize the need to perturb all frames sampled by V-MLLMs for optimal attack performance.

\textbf{Lacking generalization in perturbing video features.}  The gains from perturbation propagation are limited due to variations in how V-MLLMs extract video features, which often causes the perturbations to overfit to the features of the surrogate model's video encoder, which reduces the generalizability of perturbations. Enhancing transferability requires targeting common elements across these features. Thus, we focus on lower-level image features. The I2V attack \cite{wei2021crossmodaltransferableadversarialattacks}, which perturbs each video frame to disrupt  image features, demonstrates that using image models as surrogates can effectively generate adversarial samples for video models. Experimental results in row 3, 4 and 5 of Tab. 1 in the main content confirm that targeting image features in video frames significantly improves the transferablity of adversarial video samples. 

\textbf{Failing to integrate multimodal information.} The I2V attack was initially developed for video classification tasks and does not consider the multimodal interactions between video and text, which are crucial for comprehensive video understanding. While I2V attack achieves improved transferability, its effectiveness in video understanding tasks remains limited. Therefore, we propose using an image-based multimodal model as a surrogate, integrating multimodal interaction information into the process of generating adversarial video samples, which leads to a significant improvement in transferability, as demonstrated in the rows 3 and 6 of Tab. 1 in the main content.

In summary, we propose using I-MLLMs as surrogates to generate adversarial video samples that incorporate multimodal interactions. In addition, we introduce a perturbation propagation technique to handle different unknown frame sampling strategies. The I2V-MLLM results in Tab. 1 in the main content demonstrate the strong transferability of our method across different V-MLLMs. 

\section{Experiment \& Analysis} \label{sup_exp}

\subsection{Experiment setting}
In this section, we provide a more detailed description of the experiment setting.

\textbf{Models.} We perform our proposed approach on five I-MLLMs: BLIP-2 \cite{li2023blip2bootstrappinglanguageimagepretraining}, InstructBLIP \cite{dai2023instructblipgeneralpurposevisionlanguagemodels}, MiniGPT-4 \cite{zhu2023minigpt4enhancingvisionlanguageunderstanding}, LLaVA \cite{liu2023visualinstructiontuning} and mPLUG-owl \cite{ye2023mplug}. 
 Our proposed methods are evaluated on four different V-MLLMs: Chat-UniVi \cite{jin2024chatuniviunifiedvisualrepresentation}, LLaVA-Next-Video \cite{zhang2024videoinstructiontuningsynthetic}, VideoChat \cite{li2024videochatchatcentricvideounderstanding} and Video-LLaMA \cite{zhang2023videollamainstructiontunedaudiovisuallanguage}, each with a Vicuna-7B \cite{vicuna2023} as the LLM. 

\textbf{Propagation methods.} Direct Propagation (DP), Optical Flow-based  Propagation \cite{dosovitskiy2015flownet} (OFP), and Bidirectional Linear Interpolation Propagation \cite{dai2017deformable} (BP) are three different perturbation propagation methods. In DP, the adversarial perturbation is directly applied to each corresponding frame segment without any adjustments, simply propagating the perturbation along the frame sequence segments. OFP calculates the optical flow changes between adjacent frames and dynamically adjusts and propagates the adversarial perturbation based on the motion information between frames, facilitating a perturbation transfer that better aligns with frame-to-frame motion patterns. BP employs bidirectional linear interpolation to propagate perturbations from adjacent key frames to the intermediate frames, thereby smoothly transmitting adversarial perturbations across the entire frame sequence.

\begin{figure}[t]
  \centering
  \includegraphics[width=1\linewidth]{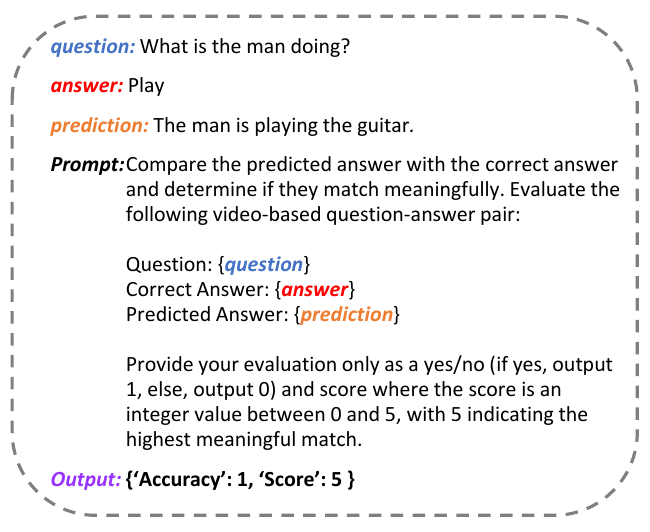} 
  \caption{An example of using GPT-4o-mini to evaluate Accuracy and GPT Score for the VideoQA task, following the methodology in \cite{maaz2024videochatgptdetailedvideounderstanding}.}
  \label{fig:videoqaprompt}
\end{figure}

\textbf{Metrics.}  We use Attack Success Rate (ASR) to evaluate the effectiveness of adversarial examples on VideoQA tasks.
It measures the percentage of successful attacks on questions the model answered correctly for clean videos. 
Answer correctness is evaluated using GPT-4o-mini \cite{openai2023gpt4omini}, which checks whether the model's prediction semantically aligns with the ground truth.
We also provide the average ASR (AASR) across all evaluated V-MLLMs. A higher ASR or AASR indicates better adversarial transferability. 
To evaluate the model's overall performance when encountering adversarial videos, we further employ GPT-assisted methods \cite{maaz2024videochatgptdetailedvideounderstanding} to assess Accuracy (Acc.) and GPT-Score. We use GPT-4o-mini~\cite{openai2023gpt4omini} as the evaluation model due to its superior performance in text understanding and its cost efficiency. Fig. \ref{fig:videoqaprompt} illustrates an example of evaluating the Zero-Shot VideoQA task with GPT-4o-mini. An adversarial attack is considered successful if the adversarial video sample causes the V-MLLM to change its response to a question from correct (Accuracy: 1) to incorrect (Accuracy: 0).

\begin{figure}[t]
  \centering
  \includegraphics[width=1\linewidth]{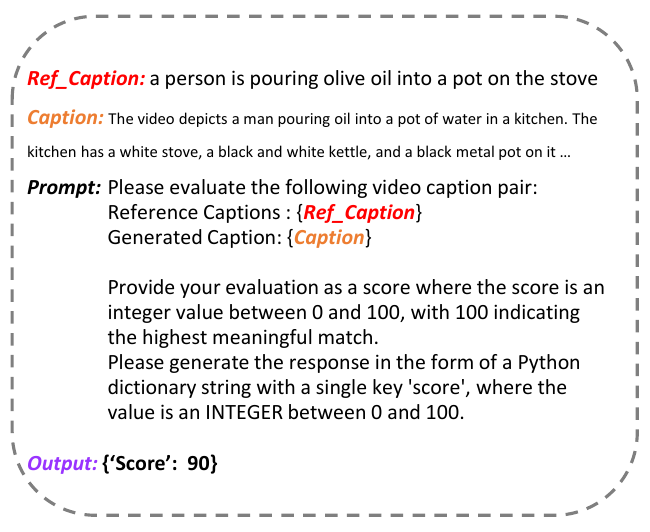} 
  \caption{An example of using GPT-4o-mini to evaluate Caption Score for the video captioning task.}
  \label{fig:eval_caption}
\end{figure}

Since traditional video captioning metrics \cite{Wang_2018_CVPR} such as BLEU, CIDEr, SPICE are not well-suited for evaluating the detailed video captions generated by V-MLLM, we employ GPT-4o-mini as the evaluation model and use the Captioning Score to assess V-MLLM's performance on the video captioning task. The Captioning Score measures the quality of the model’s predictions by assigning a relative score on a scale from 0 to 100 (see Fig. \ref{fig:eval_caption}). A higher Captioning Score indicates that the generated captions are closer to the reference captions.

\subsection{Cross-task transferability } \label{sec:exp_video_caption}
Multimodal interactions and alignments are the core components of multimodal learning, regardless of the specific task. Therefore, we conduct extensive experiments to investigate the cross-task transferability of I2V-MLLM from VideoQA to video captioning tasks.
Since traditional video captioning metrics \cite{Wang_2018_CVPR} such as BLEU, CIDEr, SPICE are not well-suited for evaluating the detailed video captions generated by V-MLLMs, we employ GPT-4o-mini as the evaluation model and use the Captioning Score to assess V-MLLM's performance on the video captioning task. The Captioning Score measures the quality of the model’s predictions by assigning a relative score on a scale from 0 to 100. A higher Captioning Score indicates that the generated captions are closer to the reference captions. We directly transfer the adversarial samples generated from the VideoQA on the MSVD-QA dataset to the video captioning tasks on the MSVD dataset. 
As shown in Tab. \ref{table:video_captioning}, I2V-MLLM causes the largest decrease in the Captioning Scores across all target models, achieving the best attack performance, even surpassing Vanilla and FMM attack methods in the white-box setting.
The white-box Vanilla attack, which achieves optimal performance in the VideoQA tasks, performs sub-optimally here because it is specifically designed for the VideoQA task, resulting in slightly weaker cross-task transferability.
This highlights that I2V-MLLM not only exhibits cross-model transferability but also demonstrates strong cross-task transferability.

\begin{table*}[t]
\centering
\renewcommand{\arraystretch}{1.2} 
\begin{tabular}{cllll}
\hline
\textbf{Attack}& \textbf{Chat-UniVi}& \textbf{LLaVA-NEXT-Video}& \textbf{VideoChat}& \textbf{Video-LLaMA}\\ \hline
Clean         & 59.64& 61.95& 54.16& 51.60\\ \hline
FMM*& 52.76{\scriptsize$\downarrow$6.88}& 58.15{\scriptsize$\downarrow$3.80}& 50.41{\scriptsize$\downarrow$3.75}& 41.02{\scriptsize$\downarrow$10.58}\\ 
Vanilla*& 25.62{\scriptsize$\downarrow$34.02}& 48.41{\scriptsize$\downarrow$13.54}& 22.20{\scriptsize$\downarrow$31.96}& 8.53{\scriptsize$\downarrow$43.07}\\ \hline
I2V& 40.49{\scriptsize$\downarrow$19.15}& 54.53{\scriptsize$\downarrow$7.42}& 46.02{\scriptsize$\downarrow$8.14}& 46.02{\scriptsize$\downarrow$5.58}\\ 
\rowcolor{gray!20}I2V-MLLM      & \textbf{22.04}{\scriptsize$\downarrow$37.60}& \textbf{38.04}{\scriptsize$\downarrow$23.91}&\textbf{6.95}{\scriptsize$\downarrow$47.21}&\textbf{3.65}{\scriptsize$\downarrow$47.95}\\ \hline
\end{tabular}
\caption{\textbf{Captioning Scores} on the MSVD dataset for video captioning tasks, ranging from 0 to 100. FMM and Vanilla are white-box attacks (marked with *) on different V-MLLMs. I2V and I2V-MLLM are black-box attacks, with their surrogate models being CLIP-L/14 and BLIP-2, respectively. {\scriptsize$\downarrow$} represents the performance drop compared to the clean video samples. The highest attack performance is highlighted in \textbf{bold}.}
\label{table:video_captioning}
\vspace{-0.25cm}
\end{table*}

\subsection{Collaboration  with textual attacks } \label{sec:exp_text_attack}

In real-world scenarios, attackers often employ a wide range of strategies to attack V-MLLMs, potentially introducing both adversarial video samples and adversarial text into the target models simultaneously. To account for this more practical and comprehensive attack scenario, we conducted additional experiments. Specifically, we applied textual attacks on the MSVD-QA dataset using the Bert-Attack method \cite{li2020bert}, modifying only a single character in each question to maximally disrupt the semantic alignment between the video and text.

As illustrated in Tab. \ref{table:text_attack}, Bert-Attack alone achieved an AASR of 37.11\%, when combined with I2V-MLLM, the AASR significantly increased to 76\%. Notably, Bert-Attack demonstrated stronger performance on Chat-UniVi and LLaVA-NEXT-Video, while our method excelled on VideoChat and Video-LLaMA. The integration of both approaches lead to consistently high ASRs across all target models, highlighting the complementary nature of textual and video-based adversarial perturbations in multimodal attack scenarios. 

These findings demonstrate that I2V-MLLM can be seamlessly integrated with textual attack methods to achieve exceptionally robust attack performance, underscoring its substantial disruptive potential in real-world applications.

\begin{table}[t]
\centering
\renewcommand{\arraystretch}{1.2} 
\resizebox{\linewidth}{!}{ 
\begin{tabular}{cccc}
\hline
\textbf{Attack}& Bert-Attack (B-A)& I2V-MLLM& I2V-MLLM + B-A\\ \hline
\textbf{Chat-UniVi}& 40.00 & 48.39& \textbf{71.13}
\\ 
\textbf{LLaVA-NEXT-Video}& 41.62 & 45.54& \textbf{66.49}
\\ 
\textbf{VideoChat}& 30.89 & 63.09 & \textbf{79.39}
\\ 
 \textbf{Video-LLaMA} & 35.92 & 74.91 & \textbf{87.51}
\\ \hline
 \textbf{AASR}& 37.11 & 57.98& \textbf{76.31}
\\ \hline
\end{tabular}
}
\caption{Attack success rates (ASR, \%)  on the MSVD-QA validation set for Zero-Shot VideoQA tasks. \textbf{Bert-Attack} from \cite{li2020bert}. }
\label{table:text_attack}

\end{table}

\begin{table*}[t]
\centering
\begin{tabular}{cccccccccc}
\hline
\multirow{2}{*}{\textbf{Target Model}}& \multicolumn{9}{c}{\textbf{$\lambda_1 : \lambda_2$}}                             \\ \cline{2-10} 
                                 & 1:1           & 1:2  & 1:3  & 1:4  & 1:5  & 2:1  & 3:1  & 4:1  & 5:1  \\ \hline
Chat-UniVi                       & \textbf{43.11} & 43.09 & 41.91 & 43.16 & 40.91 & 42.14 & 41.35 & 40.66 & 40.22 \\
LLaVA-NeXT-Video                 & \textbf{35.67} & 34.08 & 33.12 & 33.50 & 32.45 & 35.25 & 35.96 & 35.16 & 35.25 \\
VideoChat                       & \textbf{51.10} & 49.50 & 49.03 & 49.62 & 50.31 & 49.72 & 50.44 & 48.97 & 48.97 \\
Video-LLaMA                      & \textbf{64.57} & 64.11 & 62.98 & 61.99 & 62.08 & 63.56 & 61.94 & 62.17 & 62.89 \\ \hline
\textbf{AASR }                            & \textbf{48.61} & 47.69 & 46.76 & 47.07 & 46.44 & 47.67 & 47.42 & 46.74 & 46.83 \\ \hline
\end{tabular}
\caption{ASR (\%) of the I2V-MLLM attack across different \textbf{weight ratios} of the vision model loss (\(\lambda_1\)) and projector loss (\(\lambda_2\)). The highest attack performance for each target model is shown in \textbf{bold}.}

\label{table_ratio}
\end{table*}

\subsection{Ablation study}
The experiments in following sections are conducted on the MSVD-QA validation set, using BLIP-2 as the surrogate model. A higher ASR or AASR reflects better adversarial transferability.

\textbf{Influence of weights of loss functions.} We vary the weights of the \(\mathcal{L}_V\) and \(\mathcal{L}_P\) to explore their relative relationship. As shown in Tab. \ref{table_ratio}, the AASR is highest when the ratio of $\lambda_1$ to $\lambda_2$ is 1:1. Therefore, we adopt this weight ratio in our experiments. 

\begin{figure}[t]
    \centering
    \includegraphics[width=1\linewidth]{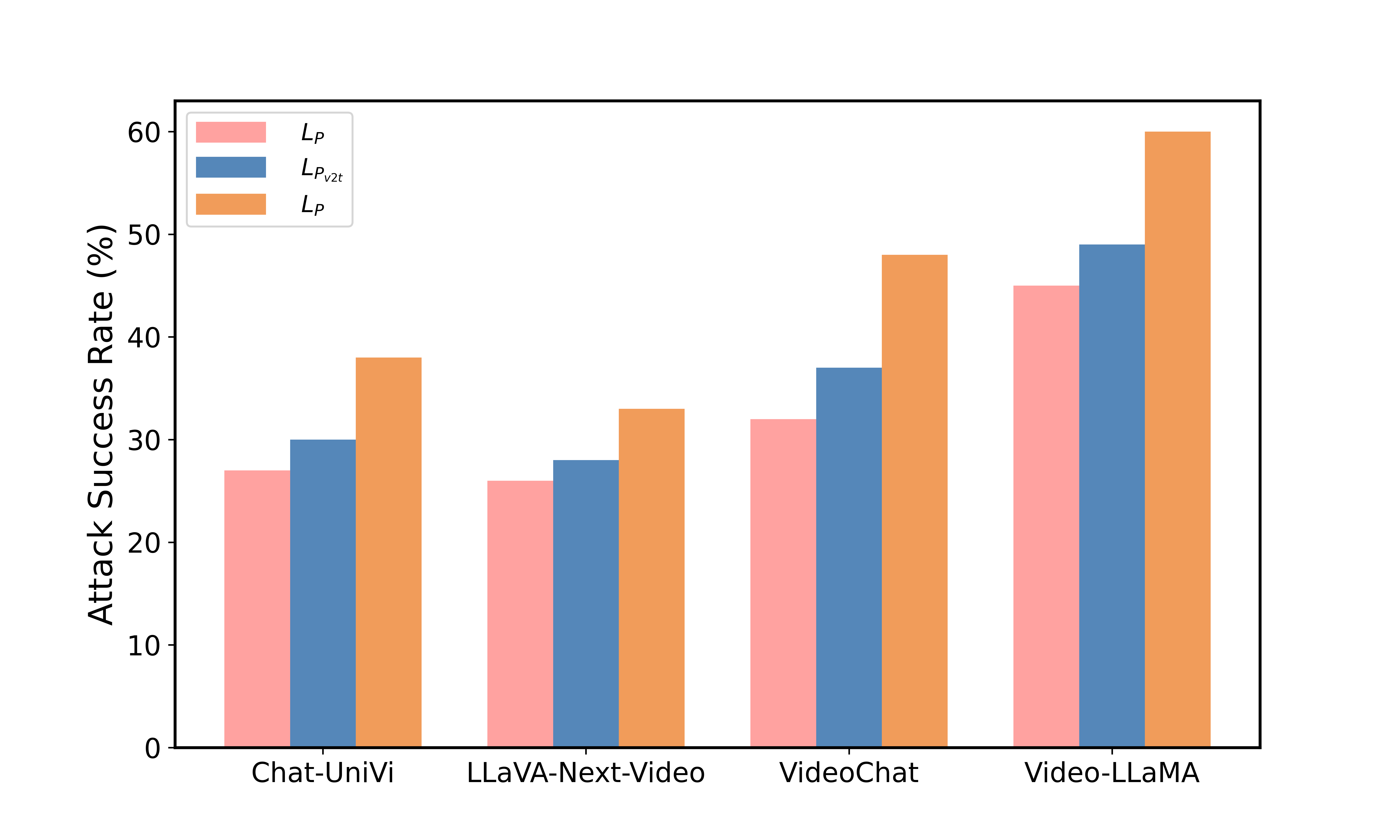}
    \caption{ASR (\%) of projector attacks in I2V-MLLM with different loss functions.}
    \label{fig:projector}
\end{figure}

\textbf{Influence of projector loss function.} We examine the influence of components of \(\mathcal{L}_{P}\). As illustrated in Fig. \ref{fig:projector}, the combination of \(\mathcal{L}_{P_{v}}\) and \(\mathcal{L}_{P_{v2t}}\) leads to an improvement in ASR, demonstrating the effectiveness of \(\mathcal{L}_{P}\) in leveraging the multimodal interactions between video and text to craft adversarial perturbations.

\textbf{Influence of vision model loss function.} In Sec. 3.3.1, Eq. (2) defines the loss function for the vision model attack, which can be further decomposed into \(\mathcal{L}_{V} = \mathcal{L}_{V}^s + \mathcal{L}_{V}^t\). 

To disrupt video-level spatial features, I2V-MLLM generates perturbations by minimizing the cosine similarity between the original and adversarial spatial features:  
\begin{equation}
\mathcal{L}_{V}^s = \sum_{i=1}^{K}\frac{\text{Cos}(F_V^{s}(X)_{i},F_V^{s}(X_{adv})_{i})}{K} ,
\label{eq:L_V^s}
\end{equation}
where \(F_V^{s}(X)_i \) and \(F_V^{s}(X_{adv})_i \) denote the \( i \)-th elements of the spatial features extracted from the original and adversarial video frames, respectively.

Similarly, to disrupt video-level temporal features, I2V-MLLM minimizes the cosine similarity between the original and adversarial temporal features:  
\begin{equation}
\mathcal{L}_{V}^t = \sum_{i=1}^{N}\frac{\text{Cos}(F_V^{t}(X)_{i},F_V^{t}(X_{adv})_{i})}{N}  ,
\label{eq:L_V^t}
\end{equation}
where \(F_V^{t}(X)_i \) and \(F_V^{t}(X_{adv})_i \) represent the \( i \)-th elements of the temporal features for the original and adversarial video frames, respectively. 

\begin{figure}[t]
    \centering
    \includegraphics[width=1\linewidth]{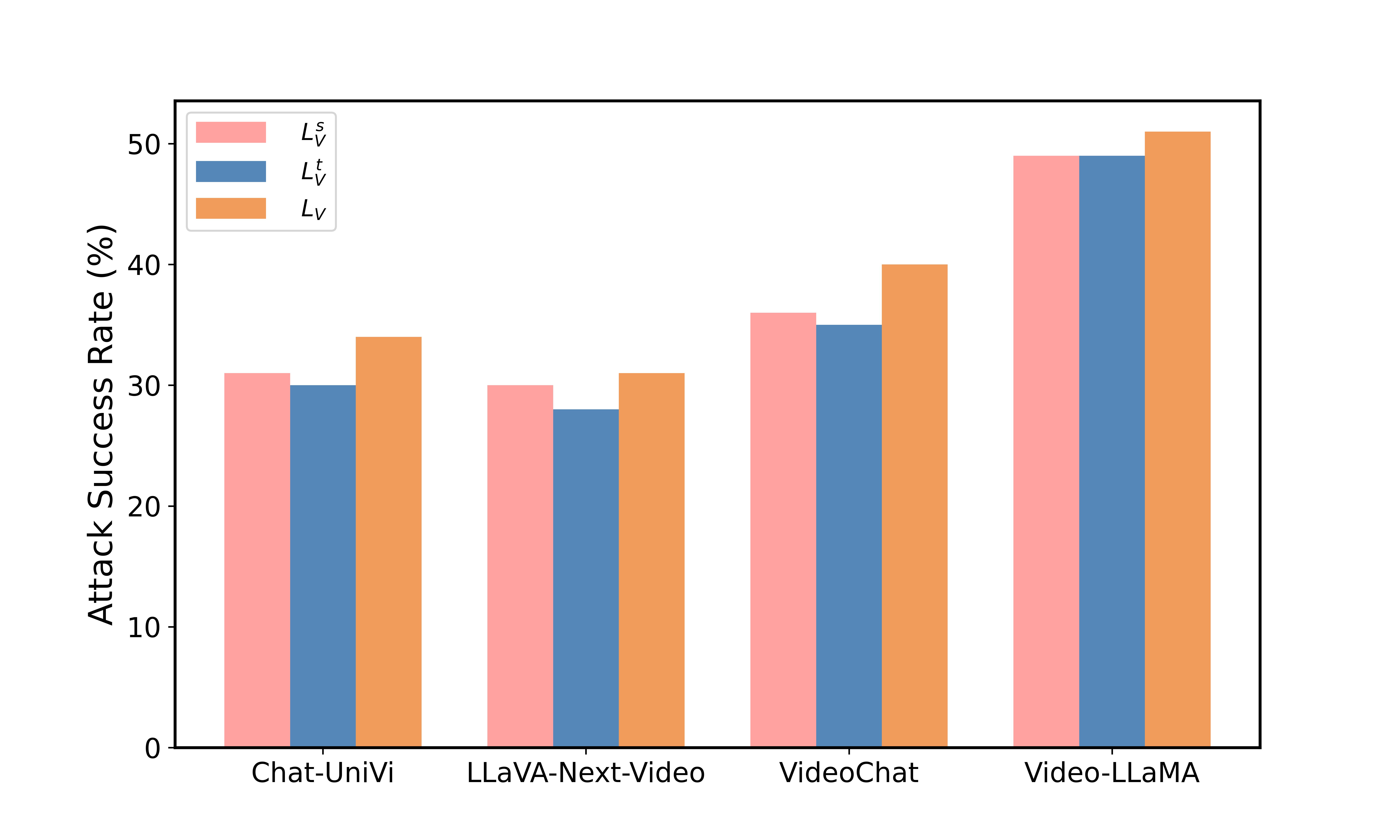}
    \caption{ASR (\%) of vision model attacks in I2V-MLLM with different \textbf{loss functions}.}
    \label{fig:vision}
\end{figure}

We analyze the individual influence of the components of \(\mathcal{L}_{V}\). As illustrated in Fig. \ref{fig:vision}, the combination of \(\mathcal{L}_{V}^s\) and \(\mathcal{L}_{V}^t\) results in an improvement in ASR, highlighting the effectiveness of \(\mathcal{L}_{V}\) in leveraging the spatiotemporal information of video samples to craft adversarial perturbations.

\begin{figure}[t]
    \centering
    \includegraphics[width=1\linewidth]{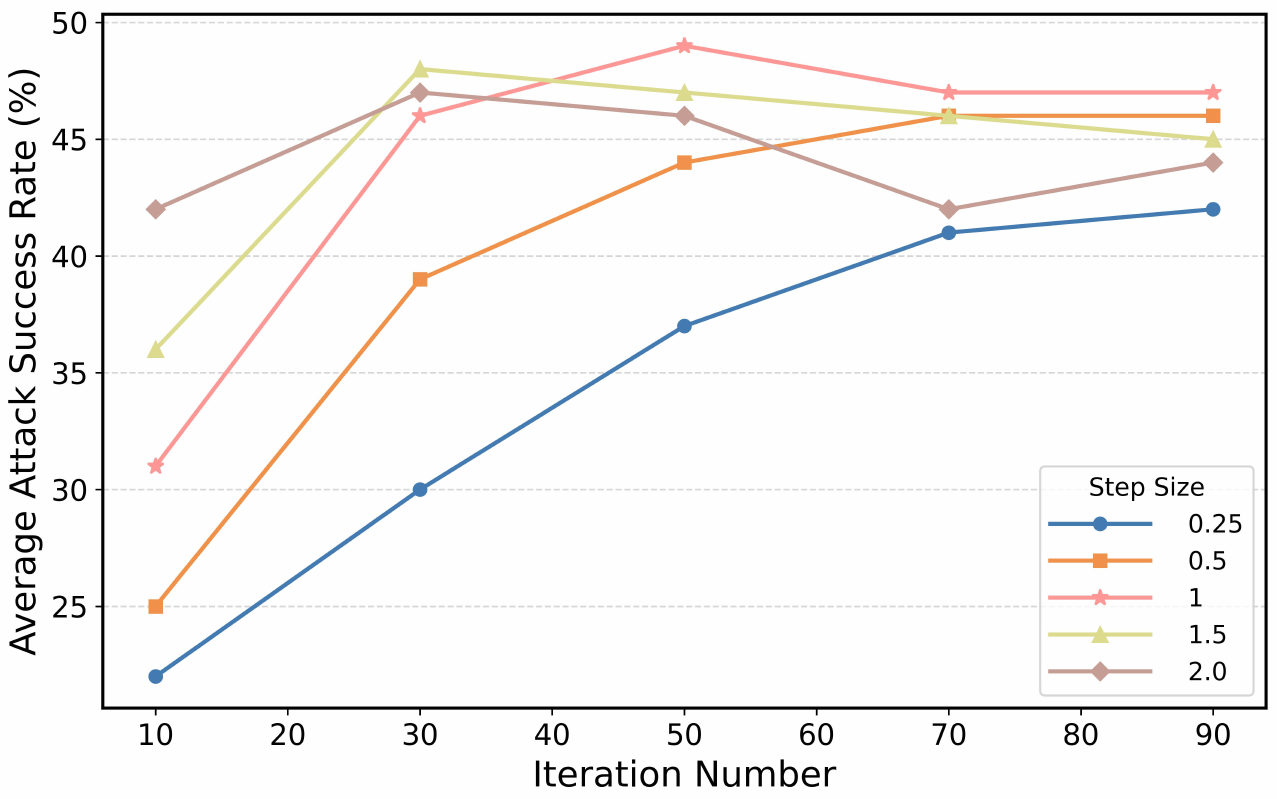}
    \caption{AASR (\%) of the I2V-MLLM attack with various step sizes and iteration numbers.}
    \label{figure 2}
    \label{fig:enter-label}
    
\end{figure}

\textbf{Influence of step size and iteration number.}  We utilize the PGD to update the perturbations, which is influenced by the step size $\alpha$ and the number of iterations $I$. 
Fig. \ref{figure 2} presents the results obtained with a key-frame ratio $\beta = 10\%$ under different step sizes and iteration numbers. 
It can be observed that as the number of iterations $I$ increases, the transferability (ASR) of adversarial examples improves, and when the iterations exceed 50, the benefits from further increases gradually diminish.
A similar pattern is observed with the step size selection.
Moderate values of $\alpha$ and $I$ yield best AASR. 
To achieve optimal performance, we adopt $\alpha = 1$ and $I = 50$ in our experiments.

\begin{figure}[t]
  \centering
  \includegraphics[trim=0 130 0 0, clip, width=1\linewidth]{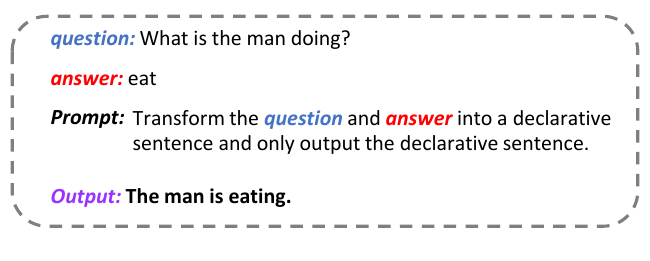} 
  \caption{An example of utilizing GPT-4o-mini to generate a caption based on the question and its corresponding answer.}
  \label{fig:captionprompt}
\end{figure}

\textbf{Influence of input text.}  When designing \( \mathcal{L}_{P_{v2t}} \) in Sec. 3.3.2, Eq. (4) of the main content, we consider two types of text inputs:  questions and captions generated from the questions and their answers. As illustrated in Fig. \ref{fig:captionprompt}, we use GPT-4o-mini to generate caption based on the question and answer. The experiment results are shown in Tab. \ref{table_text}, using captions as input yields a slightly higher AASR compared to using questions. This is because captions contain answer-related information, and the perturbations introduced during the iterations disrupt the semantic information within the answers, making it more challenging for V-MLLMs to provide responses aligned with the ground truth.

\begin{table}[t]
\centering
\renewcommand{\arraystretch}{1.2}
\begin{tabular}{ccc}
\hline
\textbf{Target Model}& \textbf{Caption} & \textbf{Question}\\ \hline
Chat-UniVi                       & \textbf{45.11}& 43.68\\
LLaVA-NeXT-Video                 & \textbf{35.67}& 35.06\\
VideoChat                       & \textbf{51.10}& 50.10\\
Video-LLaMA                      & \textbf{65.57}& 64.54\\ \hline
\textbf{AASR}                             & \textbf{49.36}& 48.34\\ \hline
\end{tabular}
\caption{This table presents the ASR (\%) across different V-MLLMs under varying \textbf{input text types}.  The highest attack performance for each target model is shown in \textbf{bold}.}
\label{table_text}
\end{table}

\textbf{Influence of perturbation bound.} To further investigate the impact of perturbation magnitude on attack performance, we tested the ASR under different perturbation bounds ($\epsilon$), including 2, 4, 8, 16, and 32. As shown Tab. \ref{table:eps}, the ASR steadily increases with higher perturbation bounds, reaching its peak when the perturbation bound is set to 32. As illustrated in Figs. 9 and 10, with the increase in perturbation bound, the perturbations become more pronounced. To balance attack performance and video vividness, we selected a perturbation bound of 16, which is also a common choice in traditional video attack methods \cite{wei2021crossmodaltransferableadversarialattacks, kim2023breakingtemporalconsistencygenerating, wang2023global}.

\begin{table}[h]
\centering
\renewcommand{\arraystretch}{1.2} 
\resizebox{\linewidth}{!}{ 
\begin{tabular}{cccccc}
\hline
\textbf{$\epsilon$}& \textbf{2}& \textbf{4}& \textbf{8}& \textbf{16}& \textbf{32}\\ \hline
Chat-UniVi& 12.96& 22.76& 32.35& \underline{48.39}& \textbf{61.32}\\ 
LLaVA-NeXT-Video& 8.14& 13.37& 24.75& \underline{45.54}& \textbf{58.37}\\ 
 VideoChat& 25.65& 32.14& 48.31& \underline{63.09}&\textbf{75.23}\\
 Video-LLaMA& 26.13& 34.27& 57.24& \underline{74.91}&\textbf{84.22}\\ \hline
 AASR& 18.22& 25.64& 40.66& \underline{57.98}&\textbf{69.79}\\ \hline
\end{tabular}
}
\caption{Attack success rate (ASR, \%) under different \textbf{perturbation bounds}. AASR represents the average ASR across all target models for each surrogate model. A
higher AASR indicates better adversarial transferability. The highest attack performance for each target model is shown in \textbf{bold} and the second-highest in \underline{underline}.}
\label{table:eps}
\end{table}

\subsection{Results on multi-faceted video understanding tasks.} \label{exp:act}
Multi-faceted video understanding tasks assess whether V-MLLMs have comprehended the content of a video by posing a range of questions about it. 
Following Maaz et al. \cite{maaz2024videochatgptdetailedvideounderstanding}, 
we use a subset of the ActivityNet-200~\cite{Heilbron_Escorcia_Ghanem_Niebles_2015} dataset
and employ GPT-4o-mini to evaluate the model's responses to adversarial examples from five perspectives: Correctness, Detail Orientation, Contextual Understanding, Temporal Understanding, and Consistency.
We compare our proposed I2V-MLLM attack with the Vanilla attack on four V-MLLMs, using clean samples as a reference. Evaluations are performed on white-box Vanilla attack and I2V-MLLM attack (using BLIP-2 as a surrogate model).

As shown in Tab. \ref{tab:video}, I2V-MLLM achieves performance comparable to the white-box Vanilla attack and even outperforms it on VideoChat and Video-LLaMA, further validating its effectiveness and transferability. 

\begin{table*}[t]
\centering
\renewcommand{\arraystretch}{1.2}
\begin{tabular}{cclllll}
\hline
\textbf{Target Model}  & \textbf{Type} & \textbf{Correct} & \textbf{Detail} & \textbf{Context} & \textbf{Temporal} & \textbf{Consistency} \\ \hline
\multirow{3}{*}{\textbf{Chat-UniVi}} 
& Clean       & 2.02          & 2.07          & 2.60          & 1.75          & 1.78 \\
& Vanilla*    & 1.33 {\scriptsize$\downarrow$0.69} & 1.44 {\scriptsize$\downarrow$0.63} & 1.81 {\scriptsize$\downarrow$0.79} & 1.36 {\scriptsize$\downarrow$0.39} & 1.32 {\scriptsize$\downarrow$0.46}\\
& I2V-MLLM    & 1.37 {\scriptsize$\downarrow$0.65} & 1.46 {\scriptsize$\downarrow$0.61} & 1.89 {\scriptsize$\downarrow$0.71} & 1.18 {\scriptsize$\downarrow$0.57} & 1.42 {\scriptsize$\downarrow$0.36}\\ \hline
\multirow{3}{*}{\textbf{LLaVA-NeXT-Video}} 
& Clean       & 2.38          & 2.54          & 2.97          & 1.97          & 1.88 \\
& Vanilla*    & 2.07 {\scriptsize$\downarrow$0.31} & 2.25 {\scriptsize$\downarrow$0.29} & 2.64 {\scriptsize$\downarrow$0.33} & 1.55 {\scriptsize$\downarrow$0.42} & 1.74 {\scriptsize$\downarrow$0.14}\\ 
& I2V-MLLM    & 2.10 {\scriptsize$\downarrow$0.28} & 2.23 {\scriptsize$\downarrow$0.31} & 2.69 {\scriptsize$\downarrow$0.28}  & 1.55 {\scriptsize$\downarrow$0.42} & 1.83 {\scriptsize$\downarrow$0.05}\\ \hline
\multirow{3}{*}{\textbf{VideoChat}} 
& Clean       & 1.87          & 2.06          & 2.44          & 1.52          & 2.00 \\
& Vanilla*    & 1.08 {\scriptsize$\downarrow$0.79} & 1.39 {\scriptsize$\downarrow$0.67} & 1.60 {\scriptsize$\downarrow$0.84} & 1.26 {\scriptsize$\downarrow$0.26} & 1.86 {\scriptsize$\downarrow$0.14}\\
& I2V-MLLM    & 1.06 {\scriptsize$\downarrow$0.81} & 1.41 {\scriptsize$\downarrow$0.65} & 1.55 {\scriptsize$\downarrow$0.89} & 1.22 {\scriptsize$\downarrow$0.30} & 1.48 {\scriptsize$\downarrow$0.52}\\ \hline
\multirow{3}{*}{\textbf{Video-LLaMA}} 
& Clean       & 1.88          & 1.89          & 2.21          & 1.64          & 1.75 \\
& Vanilla*    & 1.27 {\scriptsize$\downarrow$0.61} & 1.32 {\scriptsize$\downarrow$0.57} & 1.44 {\scriptsize$\downarrow$0.77} & 1.29 {\scriptsize$\downarrow$0.35} & 1.36 {\scriptsize$\downarrow$0.39}\\
& I2V-MLLM    & 1.26 {\scriptsize$\downarrow$0.62} & 1.33 {\scriptsize$\downarrow$0.56} & 1.42 {\scriptsize$\downarrow$0.79} & 1.34 {\scriptsize$\downarrow$0.30} & 1.25 {\scriptsize$\downarrow$0.50}\\ \hline
\end{tabular}
\caption{The results on the \textbf{ActivityNet-200} for multi-faceted video understanding tasks. All scores range from 1 to 5, with lower scores indicating better attack performance. {\scriptsize$\downarrow$} represents the score reduction compared to the clean video samples. * indicates a white-box attack.}
\label{tab:video}
\end{table*}

\subsection{Analysis}

In this section, we will discuss the consistency of intermediate features between I-MLLM and V-MLLM, as well as present more cases of successful adversarial attacks.

\begin{figure}[t]
    \centering
    \includegraphics[width=1\linewidth]{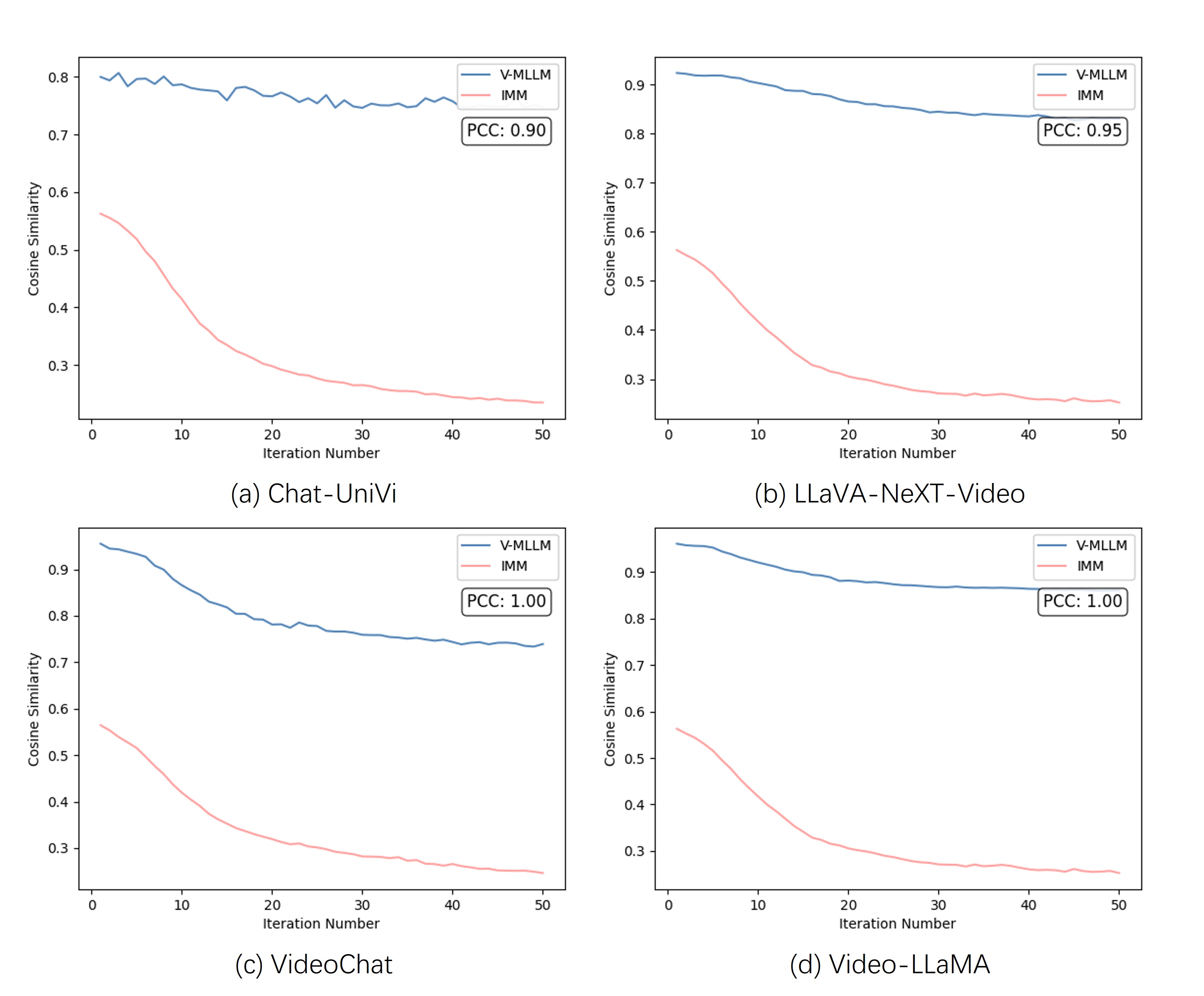}
    \caption{Pearson Correlation Coefficient (PCC) analysis between the cosine similarity trends computed from BLIP-2 and four V-MLLMs. The features of BLIP-2 are derived from vision model and projector, while the features of the V-MLLMs are obtained from the video encoders and the LLMs.}
    \label{fig:PCC}
\end{figure}

\textbf{Discussion.} To experimentally validate the effectiveness of Eq (6) in the main content, we analyze how the cosine similarity between adversarial and benign features in I-MLLM/V-MLLMs evolves as the iteration number increases. The Pearson Correlation Coefficient (PCC) \cite{Schmee_1986} is used to quantify the linear correlation between cosine similarity trends computed from both I-MLLM and V-MLLMs.  Fig. \ref{fig:PCC} presents the PCC analysis of these trends, using BLIP-2 and four different V-MLLMs. As shown, all PCC values exceed 0.90, indicating a strong positive linear relationship between the directional changes of intermediate features in I-MLLM and V-MLLM. This suggests that perturbations in I-MLLM’s image features can effectively disrupt the intermediate features of video samples in V-MLLMs. Notably, the PCC values between BLIP-2 and VideoChat, as well as Video-LLaMA, are exactly 1, which aligns with the highest ASR values observed for these models in Tab. 2 in the main content. The slightly lower PCC values with Chat-UniVi and LLaVA-NeXT-Video correspond to the lower ASR values, demonstrating that a higher PCC between I-MLLM and V-MLLMs indicates better adversarial transferability.

\begin{figure}[t]
    \centering
    \includegraphics[width=1\linewidth]{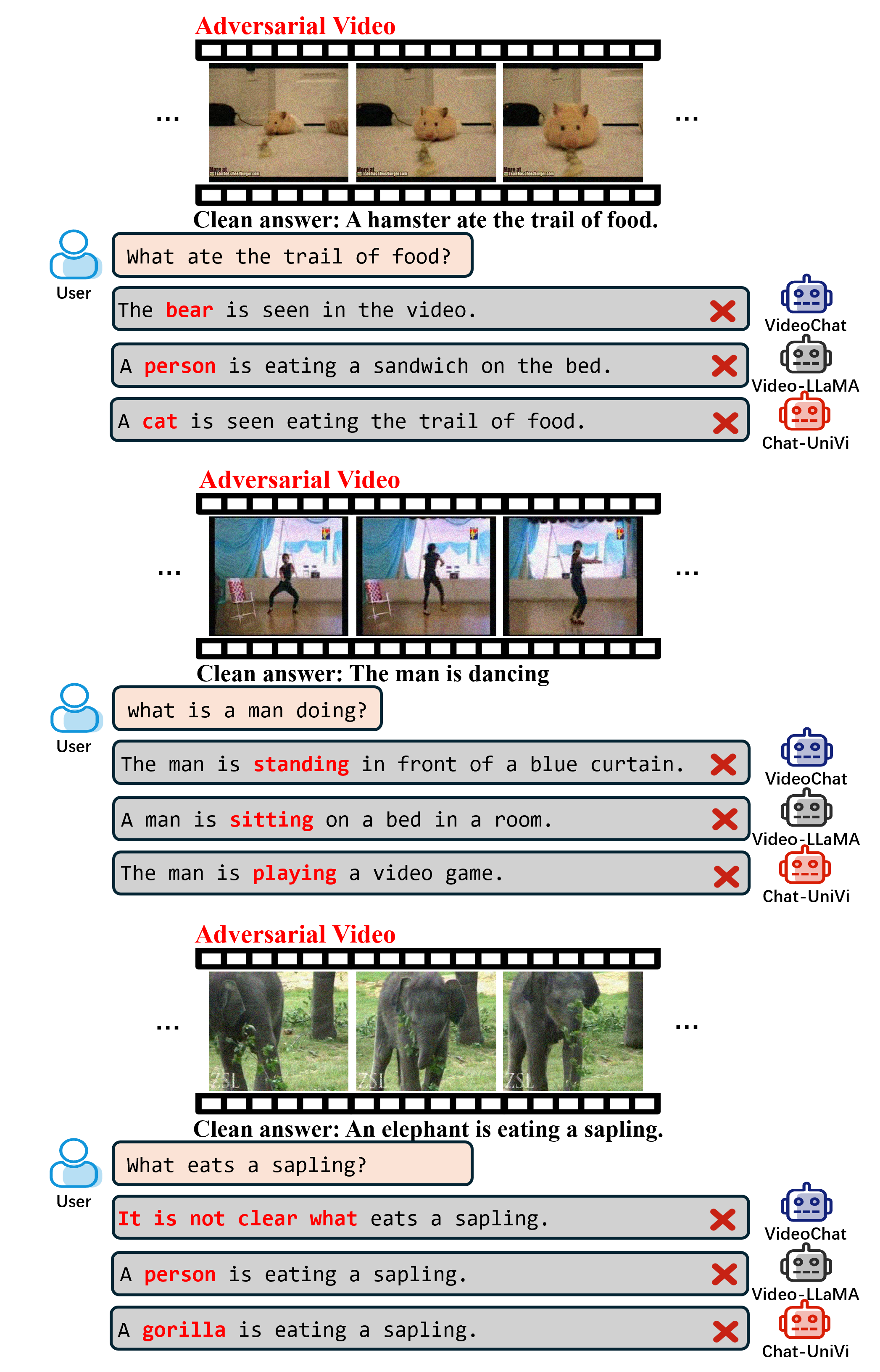}
    \caption{The adversarial video samples for VideoQA tasks are based on MSVD-QA, where the surrogate model is BLIP-2 and the target V-MLLMs are VideoChat, Video-LLaMA, and Chat-UniVi. The clean answers are the responses generated by Chat-UniVi on clean video samples. Red crosses indicate that the responses generated by V-MLLMs do not semantically align with the expected clean answers.}
    \label{fig:more_case}
\end{figure}

\textbf{Case study.} As shown in Fig. \ref{fig:more_case}, adversarial video samples generated from the I2V-MLLM attack cause different V-MLLMs to produce responses that differ significantly from the clean answers, demonstrating that our method effectively misleads V-MLLMs and disrupts their ability to accurately interpret the video content.

\begin{algorithm}[t]
\floatname{algorithm}{Algorithm}
    \caption{I2V-MLLM Attack} 
    \label{algorithm:1}
\setcounter{algorithm}{0}
{\bf Input:} A video sample $V$, caption set $T$.\\
{\bf Parameters:} Step size $\alpha$, iteration number $I$, perturbation budget $\epsilon$, key-frame ratio $\beta$, loss function weights $\lambda_1$, $\lambda_2$.\\
{\bf Output:} Adversarial sample $V_{adv}$.
\begin{algorithmic}[1]
\STATE \textbf{// key-frame Selection}
\STATE Split video $V$ into $K$ clips using key-frame ratio $\beta$, extract the first frame $x_k$ from each clip $v^k$, forming key-frames $X = \{x^1, x^2, \ldots, x^K\}$

\STATE \textbf{// Perturbation Optimization}
\STATE Initialize $X_{adv} = X  + \delta_0, \delta_0 \in U(-\epsilon, \epsilon)$
\STATE Get loss function \(\mathcal{L}_{total}\).
\FOR {\(i = 0, ..., I - 1\)} 
    \STATE Calculate gradient for adversarial frames:
    \STATE\quad \(g = \nabla_{X_{adv}}\mathcal{L}_{total}\)
    \STATE Update \(\delta_{i + 1}\) with gradient descent:
    \STATE  \quad $\delta_{i + 1} = \delta_{i} - \alpha \cdot sign(g)$ 
    \STATE Project \(X_{adv}\) to $\epsilon$-ball of \(X\):
    \STATE \quad \(X_{adv} = clip_{X, \epsilon}( X + \delta_{i+1})\) 
\ENDFOR
\STATE \textbf{// Perturbation Propagation}
\FOR {$k = 1, 2, ..., K$}
    \STATE Propagate $\delta^k_I$ to the video clip $v^k$, yielding $\delta'^k$
\ENDFOR
\STATE \textbf{// Construct the adversarial video:}
\[
V_{adv} = \{v^1 + \delta'^1, v^2 + \delta'^2, \ldots, v^K + \delta'^K\}
\]
\STATE \textbf{Return} the adversarial video sample $V_{adv}$
\end{algorithmic} 
\end{algorithm}

\section{Algorithm} \label{sup:algorithm}
The complete I2V-MLLM Attack process is described in Algorithm \ref{algorithm:1}.

\begin{figure*}[t]
    \centering
    \includegraphics[width=1\linewidth]{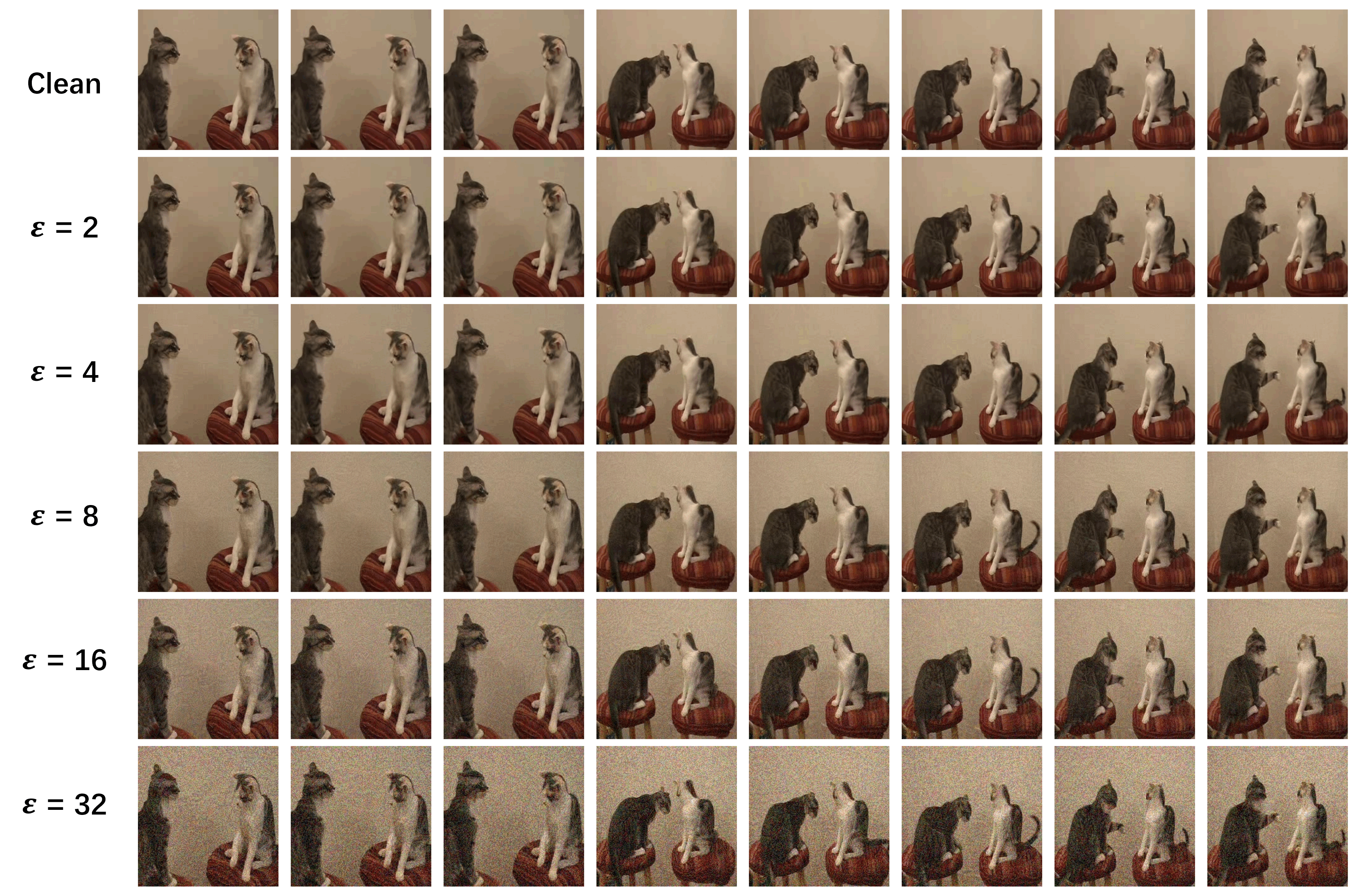}
    \caption{Comparison of adversarial video frames generated by I2V-MLLM under different perturbation bounds. }
    \label{fig:case1}
\end{figure*}

\begin{figure*}[t]
    \centering
    \includegraphics[width=1\linewidth]{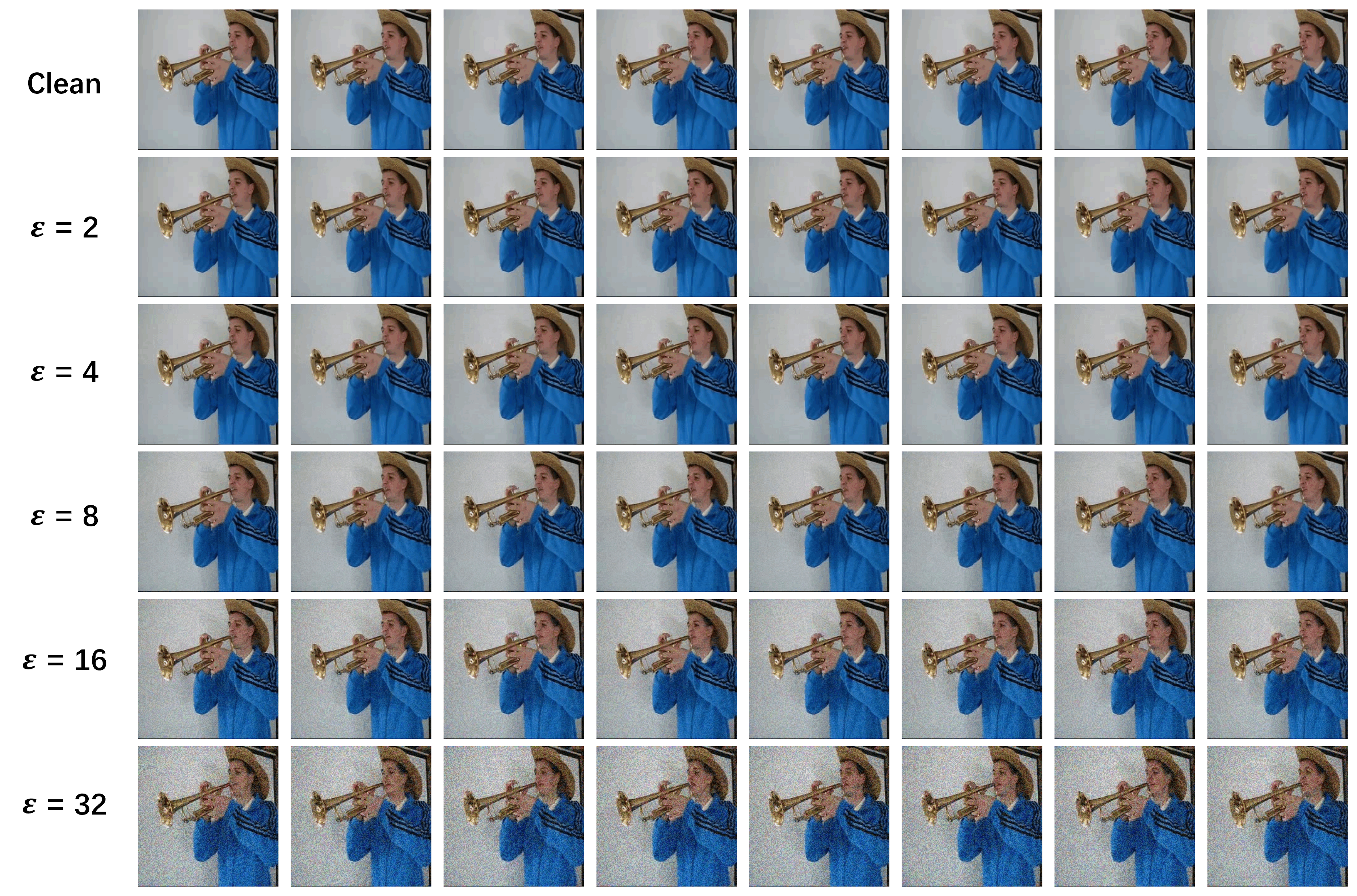}
    \caption{Comparison of adversarial video frames generated by I2V-MLLM under different perturbation bounds. }
    \label{fig:case2}
\end{figure*}
